\pdfoutput=1

\documentclass[11pt]{article}

\usepackage[]{acl}

\usepackage{times}
\usepackage{latexsym}

\usepackage[T1]{fontenc}

\usepackage[utf8]{inputenc}

\usepackage{microtype}
\usepackage{booktabs}
\usepackage{algorithm}
\usepackage{algorithmic}
\usepackage{microtype}
\usepackage{hyperref}
\usepackage{multirow}
\usepackage{amsmath}
\usepackage{amssymb}
\usepackage{mathtools,caption,subcaption,svg,stfloats}
\usepackage{enumitem}
\usepackage{newfloat}
\usepackage{listings}
\usepackage{color, colortbl}
\usepackage{tabularx}
\usepackage[export]{adjustbox}
\usepackage{wrapfig}

\usepackage{latexsym}

\title{Leveraging Machine-Generated Rationales to Facilitate Social Meaning Detection in Conversations}

\author{Ritam Dutt, Zhen Wu, Kelly Shi, Divyanshu Sheth,\\
\textbf{Prakhar Gupta, Carolyn Penstein Ros\'{e}} \\
\\
Language Technologies Institute, Carnegie Mellon University \\
\texttt{\{rdutt,zhenwu,jiaxins1,dasheth,prakharg,cprose\}@cs.cmu.edu}\\
}

\begin{document}
\maketitle

\begin{abstract}

We present a generalizable classification approach that leverages Large Language Models (LLMs) to facilitate the detection of implicitly encoded social meaning in conversations. We design a multi-faceted prompt to extract a textual explanation of the reasoning that connects visible cues to underlying social meanings.  These extracted explanations or rationales serve as augmentations to the conversational text to facilitate dialogue understanding and transfer. Our empirical results over 2,340 experimental settings demonstrate the significant positive impact of adding these rationales. Our findings hold true for in-domain classification, zero-shot, and few-shot domain transfer for two different social meaning detection tasks, each spanning two different corpora. 
\end{abstract}

\section{Introduction}
\begin{quote}
    ``All the world's a stage, and all the men and women merely players.'' \cite{shakespeare:asyoulikeit}
\end{quote}

Beyond content focused areas of Natural Language Processing (NLP), the past two decades have witnessed a surge of interest in modeling language from a social perspective \cite{nguyen2016computational}. According to  sociologist Erving Goffman \cite{goffman2002front} language conveys two forms of ``social meaning'', namely, one that is \emph{given} or intentional, and one that is \emph{given off} or unintentional, often thought of as ``reading between the lines''.

The former  embodies the idea of linguistic agency, the deliberate choices people make to protect their identity \cite{gee2014introduction} or to accomplish social goals \cite{martin2003working}. The latter encompasses involuntary cues that signal dispositions, like personality \cite{mairesse2006words,moreno2021can}, attitude \cite{martin2003language}, or emotion \cite{hazarika2018icon}, or psychological conditions, like mental illness  \cite{kayi2017predictive,alqahtani2022quantitative}, 

Since social meaning is subtly encoded, traditional classification models often over-fit to context-specific linguistic elements that correlate with these subtle cues within context. Consequently, this makes transfer to unseen domains especially challenging. For example, the same strategy to resist being persuaded would manifest in different ways depending on whether one is negotiating the price of a commodity, or one is hesitating donating to charity \cite{dutt-etal-2021-resper}. In this work, we propose a generalizable framework that leverages Large Language Models (LLMs) for detecting different kinds of social meaning in conversations.

\begin{figure}
    \centering
    \includegraphics[width=1.05\linewidth]{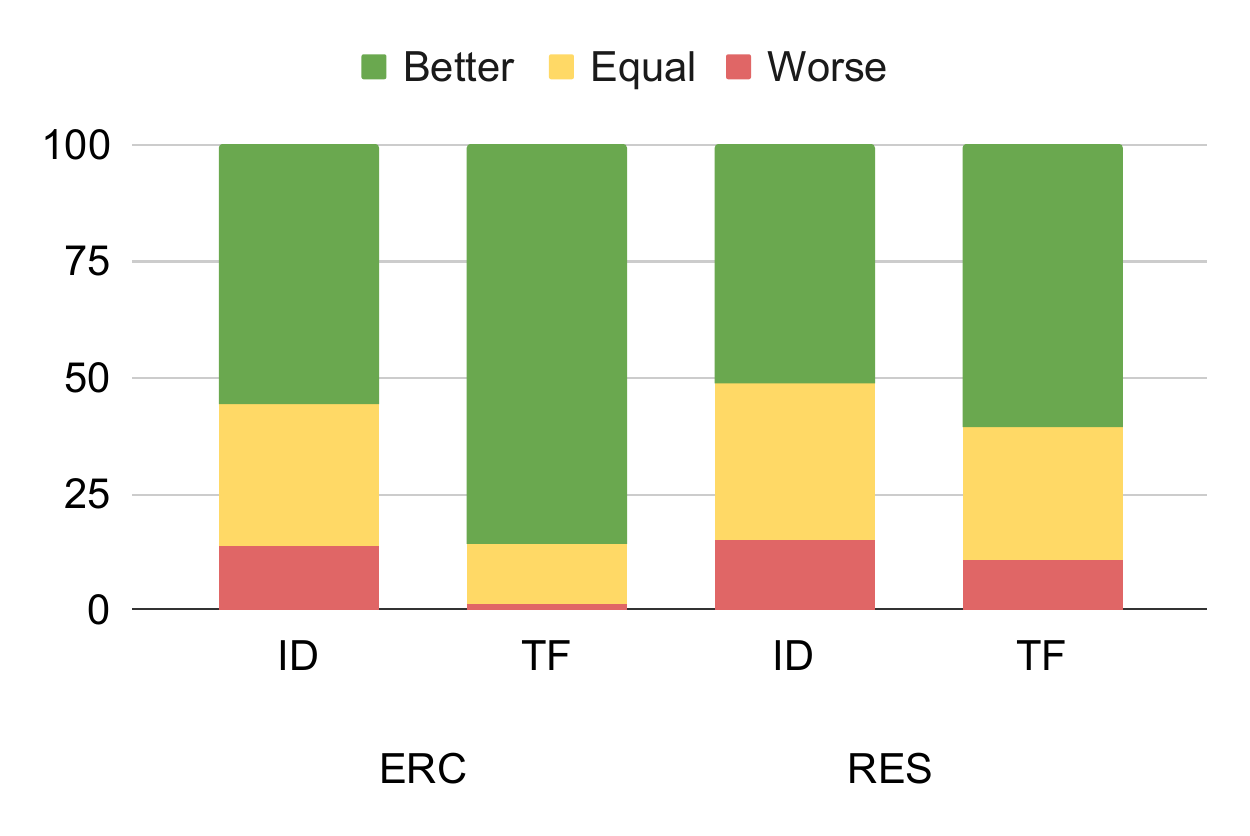}
    \caption{Fraction of cases where the classification performance was significantly better, same, or worse, when rationales were augmented, for two different tasks, i.e. detecting resisting strategies (RES) and recognizing emotions (ERC) and for two settings i.e., in-domain (ID) and transfer (TF). }
    \label{fig:overview-results}
\end{figure}

We systematically investigate the generation of ``rationales'' by LLMs, that are designed to break through the opaque surface form of the conversation's text and make the social cues more transparent. While rationales have been utilized previously, to facilitate reasoning \cite{what-makes-it-ok-2023, zelikman2022star}, or to explain model predictions \cite{wiegreffe-etal-2021-measuring}, we use rationales to refer to the elicited social meaning, i.e., why and how an utterance was conveyed in dialogue.  

Our empirical study examines the role of augmenting rationales for two specific social meaning detection tasks: (i) Resistance Strategies (RES), which aligns with intentional and purposeful communication, and (ii) Emotion Recognition (ERC), which is characterized by habitual and subconscious responses. For each of these tasks, the evaluation is conducted over two separate corpora (different domains), but the same social meaning detection task. And thus we present results both for the in-domain (ID) and transfer (TF) settings.  We illustrate in Figure~\ref{fig:overview-results} that baseline models performed significantly worse than their rationale-augmented counterparts for both tasks and settings. 
Our contributions are as follows :


\begin{itemize}

\item We investigate the role of rationales for conveying social meaning by making explicit the subtle cues implicitly encoded during a conversation. 

\item We design a multi-faceted prompting framework, grounded in sociolinguistic theory, to generate  rationales of high quality.

\item We demonstrate the positive impact of adding rationales for two social meaning detection tasks across several models.

\item We observe that rationales lead to greater performance gains in a cross-domain setting, especially in low data regimes, thereby highlighting the generalizability of our approach.

\end{itemize}

We provide the datasets augmented with rationales and code as public resources to encourage future research, especially for the purpose of developing open-source solutions that achieve the same functionality as the proprietary LLMs that perform best in our studies. We also make our code and data publicly available here \footnote{\url{https://github.com/ShoRit/RATDIAL}}.

\section{Related Work}

\subsection{Social Meaning in NLP}

Social meaning is the signaling people do during interactions to maintain positioning in terms of identity and relationship (e.g., practices of signaling are defined in detail in \citet{gee2014introduction}, with additional operationalizations in \citet{martin2003language} and \citet{meyerhoff2019pursuit}).  It encompasses both the linguistic agency and goals of the speaker (``the explicit") as well as their personal characteristics and dispositions (``the implicit") \cite{goffman2002front}. 

While originally defined in the context of socio-linguistics, the term ``social meaning'' been heavily used in the computational linguistics community. It can refer to different interactional styles \cite{jurafsky2009extracting}, or the social background and identity of a user that can be predicated from linguistic variation \cite{nguyen2021learning}, or the meaning that emerges through human interaction on social media in the form of emotion, sarcasm, irony and the like \cite{zhang-abdul-mageed-2022-improving}.  

Given the myraid definitions of the same, we adopt ``social meaning'' as an umbrella term to refer to tasks that infer the intentions of the users or their characteristics in a social setting. Specifically, in this work we focus on two social meaning detection tasks, namely the strategies employed by an individual to resist persuasion (RES) or the emotions expressed during a conversation (ERC). 


\subsection{Generalization in Dialogue}

Generalization in the context of dialogue tasks is a challenge because the interaction is typically organized around a task rather than the presentation of information, has multiple loci of control, and so much is implicit in it.   \citet{mehri2022towards} provides an outline of different kinds of generalization imperative for dialogue. These include (i) new inputs arising from covariate shift or stylistic variation \cite{khosla2022covariate-shift}, (ii) new problems in dialogue modeling such as evaluation and response generation \cite{fewshotoz} (iii) new outputs and schemas corresponding to out-of-domain shift \cite{clinic} and (iv) new tasks like controlled generation or fact verification \cite{gupta-etal-2022-instructdial}. 

Politeness is a good example of a social meaning where work on generalizability has been frequent, and in fact, the theory itself was designed with the intention of generalizability \cite{brown1987politeness}.  This particular theory has been operationalized computationally using a wide variety of approaches as the field has evolved \cite{danescu2013computational,li2020studying, dutt-etal-2020-keeping}.  In practice, generalizability is still challenging \cite{khan2023transformer}, because the features that garner the most influence within trained models tend to be domain-specific or target only the relatively infrequent, strongly overt forms of politeness.
Another notable work on transfer for social meaning detection is that of \citet{HAZARIKA20211} where they designed a hierarchical dialogue model, pretrained on multi-turn conversations and subsequently adapted for emotion classification.

\subsection{Rationales in NLP}

In the context of NLP, the term ``rationales'' has long been used to refer to \textit{textual explanations}, either generated by machines or humans. Rationales serve a wide variety of purposes such as facilitating commonsense and social reasoning \cite{zelikman2022star, majumder2022knowledge}, explaining the predictions of neural models \cite{wiegreffe-etal-2021-measuring, jayaram-allaway-2021-human, zaidan2007using}, and even assisting humans in their tasks \cite{das2020leveraging, joshi2023machine}.

Recent research has demonstrated the efficacy of LLMs in generating step-by-step explanations or rationales \cite{Gurrapu_2023} that can be harnessed to bolster downstream task performance \cite{what-makes-it-ok-2023, wei2022chain, zelikman2022star}. Rationales have also contributed to the OOD generalization of models \cite{majumder2022knowledge, xiong2023rationaleenhanced, joshi-etal-2022-er}.


Building upon this foundation, we frame rationales as the elicited verbalization of social meaning in a conversation; they make explicit the underlying social signals and helps overcome some limitations of static text like omission of communicative intent \cite{sap-etal-2022-neural}. 
We make a distinction from prior works on social reasoning \cite{what-makes-it-ok-2023, sap-etal-2020-social} which uses rationales as means of contextualizing a task with pre-conceived social norms, whereas we use rationales to elicit the implicit intentions and assumptions of the speaker.

\section{Prompting Framework}

\begin{figure*}
    \centering
    \includegraphics[width=\linewidth]{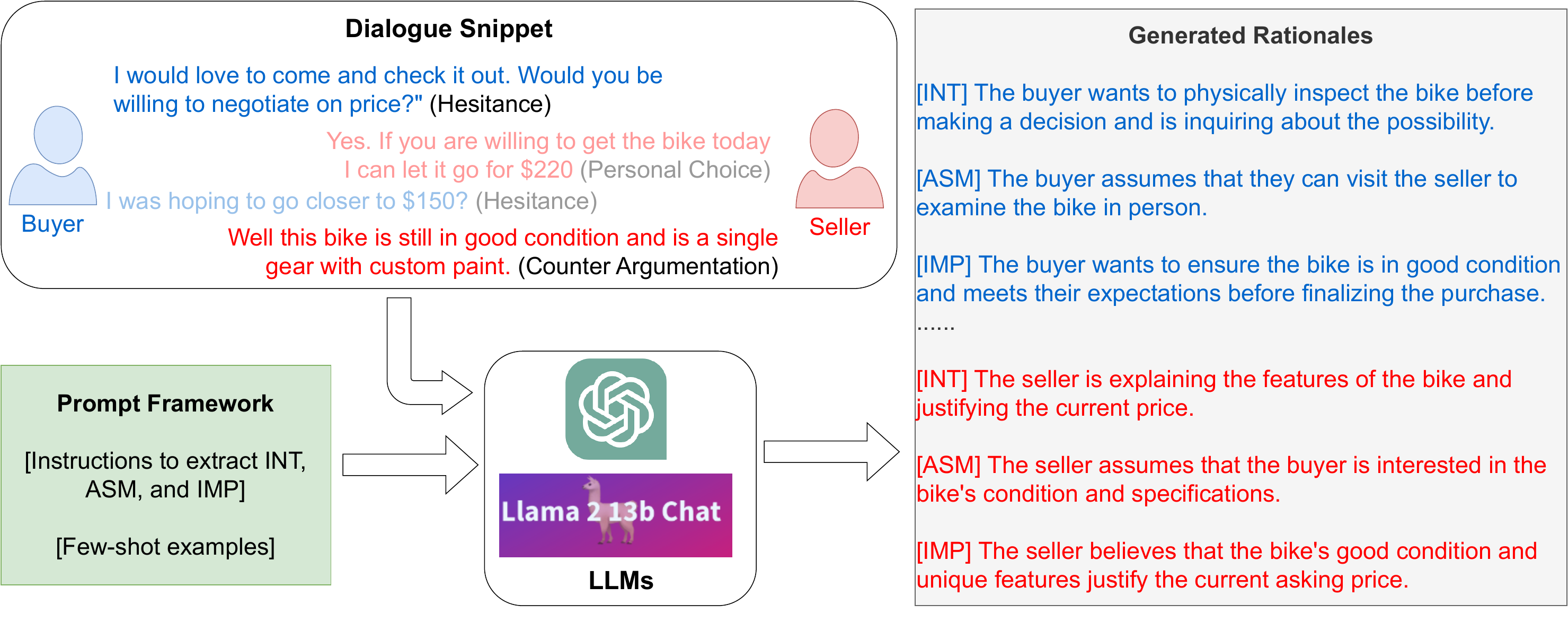}
    \caption{We present the prompting framework employed in this work to generate rationales that are subsequently used for dialogue understanding and transfer using pre-existing LLMs such as GPT-3.5-turbo and LLama-2 variants. We feed in the prompt (green box on the left) for a given dialogue to generate the speaker's intentions (INT), assumptions (ASM), and the underlying implicit information (IMP) (gray box in the right). For lack of space we showcase the generated rationales only for the first (in blue) and last utterance(in red). 
    }
    \label{fig:prompt-design}
\end{figure*}



In this section, we propose a prompting framework to generate rationales that can capture the underlying social meaning and assess their validity.  We showcase our prompting framework in Figure \ref{fig:prompt-design}. 

\subsection{Prompt Design Motivation}

The design for our prompts was grounded in \citet{goffman2002front}'s notion of social meaning in language; both intentional and accidental. Dialogue understanding relies on pragmatic reasoning to recognize subtle clues that are \emph{implicit} or obscured by the surface form, often thought of as ``reading between the lines''.  Accurate interpretation also includes what \emph{assumptions} underlie the choices made by the speaker, and choices that may reveal aspects of the speaker's \emph{intentions}. 

Motivated by this conceptualization of social meaning, we prompt the LLM to generate rationales that adhere to the speaker's intention, their underlying assumptions, and any implicit information present in the conversation (henceforth referred to as INT, ASM, and IMP respectively).  We briefly describe the three different rationale types below.


\noindent  \textbf{(i) Intention (INT)} refers to the underlying purpose or goal that a speaker seeks to achieve or communicate. It captures the deliberate messages conveyed in the dialogue.

\noindent \textbf{(ii) Assumptions (ASM)} refer to the biases or presumptions that the speaker holds. They often reflect the speaker's background, experiences, societal norms, and unacknowledged biases.

\noindent \textbf{(iii) Implicit Information (IMP)} encompasses the information that, while not overtly expressed, is inferred or understood within the context of the conversation. It offers essential cues about the conversation and its nuances.



\subsection{Structured Prompting} 

We adopt a ``structured prompting'' approach inspired by recent work that crafts prompts in a code-like-manner, such as utilizing python's dictionary data structure \cite{jung-etal-2023-enhancing, madaan-etal-2022-language} or as pseudo-code \cite{mishra2023prompting}. In our case, the prompt had the following four components, namely  (i) description of the high-level task, i.e., analysis of social meaning in dialogue, (ii) instructions that outline the generation of rationales, i.e., the elicitation of  speaker's intention, assumptions, and implicit information (i.e. INT, ASM, and IMP) in a procedural manner,  (iii) an output template that specifies the format in which the response is to be structured, and (iv) examples of input-output pairs consistent with the template.


We observed that prompting LLMs to generate all three rationales (INT, ASM, and IMP) together facilitated instruction following. Hence we term our approach as ``multi-faceted prompting''. These rationales were positioned as augmentations to the conversational text for two downstream social meaning detection tasks.  We provide examples of prompts for the two tasks in Tables \ref{table:erc-prompt} and  \ref{table:res-prompt} in the Appendix.

\subsection{Dialogue Context \& In-Context Examples}

Even for humans, understanding an individual utterance is challenging in absence of the situated dialogue context. Consequently, for our prompting framework, we provide each utterance with the corresponding dialogue history in the form of the five preceding utterances. During  development process, we experimented with different sized context windows, and five turns achieved the best result.



Furthermore, since LLMs are effective few-shot learners \cite{wei2022emergent}, we also provide the prompts with a few in-context examples to improve response generation. These in-context examples were generated using GPT4
\cite{gpt4}.

\subsection{Validity of Generated Rationales}

To assess the quality of the generated rationales, we prompted two prevalent pre-trained LLMs in contemporary NLP research; GPT-3.5-turbo-16k or ChatGPT\footnote{https://platform.openai.com/docs/models/gpt-3-5} and the Llama2-13B-Chat \cite{touvron2023llama}   
to generate rationales. We sampled 20 instances from each dataset (80 in total) to compare the generation quality of the models. The  assessment, which involved choosing the  output with a higher quality, was carried out by three graduate students proficient in  English. The results of our experiments present in Table \ref{tab: chatgpt-vs-llama-ratgen} of the Appendix showcases that annotators prefer the ChatGPT model 75\% of the times, and hence we adopted it as the LLM of our choice for subsequent experiments. 

Furthermore, to measure the generation quality, we provided two annotators with the aforementioned 80 rationales and asked them to score how grammatical, relevant, and factual the rationales are on a Likert scale (from 1-5, with 5 being the best), in accordance with past work on generation. We describe the details of the annotation process, and qualitative analysis in the Appendix \ref{sec: annotation analysis}.  

Overall, we observe an average score of 5.0, 4.6, and 4.8 for grammaticality, relevance, and factuality respectively.  We also compute the inter-rater reliability scores (IRR) for these 3 dimensions  using the multi-item agreement measure of \citet{lindell1999revised} and observe strong agreement scores for all three criteria: grammaticality (0.99), relevance (0.95), and factuality (0.96). Our qualitative analysis reveals that the rationales generated are of high quality and thus we use them for our downstream tasks of social meaning detection. We present the mean value of the Likert scores for the annotators on the 4 datasets in Table \ref{tab: rat-quality}.

\begin{table}[h]
\centering

\caption{We present here the manual evaluation scores (ranging from  1 to 5 with 5 being the best) for ChatGPT-generated rationales on the used datasets.}
\resizebox{\linewidth}{!}{
\begin{tabular}{lccc}
\toprule
\textbf{Dataset} & \textbf{Grammar} & \textbf{Relevance} & \textbf{Factuality} \\ \midrule
Friends & 5.00      & 4.55      & 4.75       \\ 
IEMOCAP & 4.98   & 4.92     & 4.34      \\ 
P4G     & 5.00       & 4.52     & 4.92      \\ 
CB      & 5.00       & 4.55      & 5.00          \\ \bottomrule
\end{tabular}}

\label{tab: rat-quality}
\end{table}

\section{Experimental Setup}


\begin{figure*}[]
\centering
\subfloat[Label Distribution in the emotion datasets]{\includegraphics[width=0.43\linewidth]{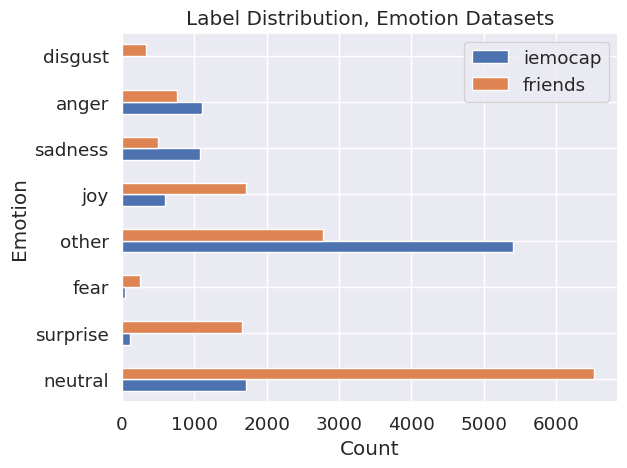}}
\subfloat[Label Distribution for the resisting strategies datasets]{\includegraphics[width=0.53\linewidth]{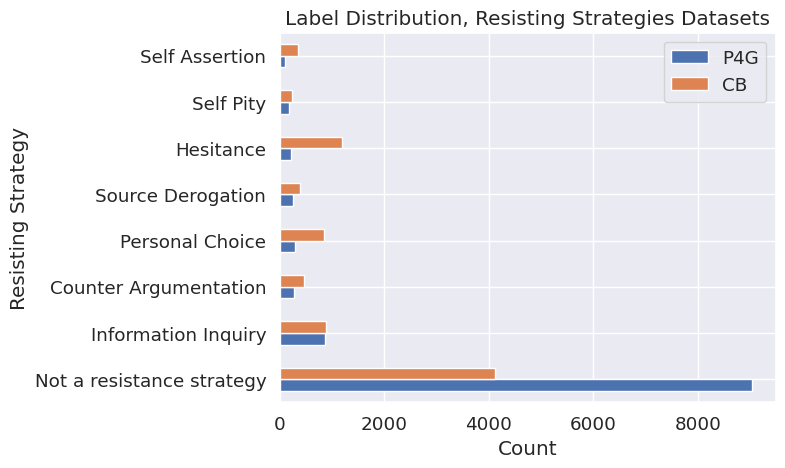}}
\vspace{-0.1cm}
\caption{We present here the label distribution for the emotion recognition and the resisting strategies datasets.}
\label{fig: datasets-label-distribution}
\end{figure*}

\subsection{Datasets}

We explore two social meaning detection tasks, namely emotion recognition in conversations, or ERC \cite{hazarika2018icon, HAZARIKA20211}, and resisting strategies detection, or RES \cite{dutt-etal-2021-resper}. We formulate both ERC and RES as utterance classification tasks, i.e., we categorize an utterance into one of several labels (8 for both ERC and RES), given its corresponding conversational context. Each task is realized via two representative datasets namely ``Friends'' \cite{hsu-etal-2018-emotionlines} and ``IEMOCAP'' \cite{iemocap} for ERC and the modified variants of the ``P4G''  and ``CB'' datasets introduced in \citet{dutt-etal-2021-resper} for RES. 

For each task, the corresponding datasets (IEMOCAP and Friends for ERC, and P4G and CB for RES) operated over the same set of labels, but they exhibit different distributions (see Figure \ref{fig: datasets-label-distribution}). Thus the two datasets for both tasks exhibit a natural covariate shift, thus making them prime candidates to investigate transfer. Furthermore, for RES, although the meaning of a given strategy remains invariant across domains, their semantic interpretation depends on the context, e.g., skepticism towards the charity in P4G vs criticism of the product in CB constitutes the same resisting strategy ``Source Derogation''.

We provide a definition for each of the eight emotions and resisting strategies along with examples for RES and ERC in Table \ref{tab:res-framework} and Table \ref{tab:erc-framework} of the Appendix respectively. We also note the fraction of instances for which the generated rationales were valid. We assess validity based on whether the response was a non-null string, had the appropriate speaker as its subject, and had information of all three rationales (i.e. INT, ASM, and IMP). We observe that valid generations account for $\approx 95\%$ of P4G, IEMOCAP and Friends .


\begin{figure*}[h]
    \centering
    \includegraphics[width=0.6\linewidth]{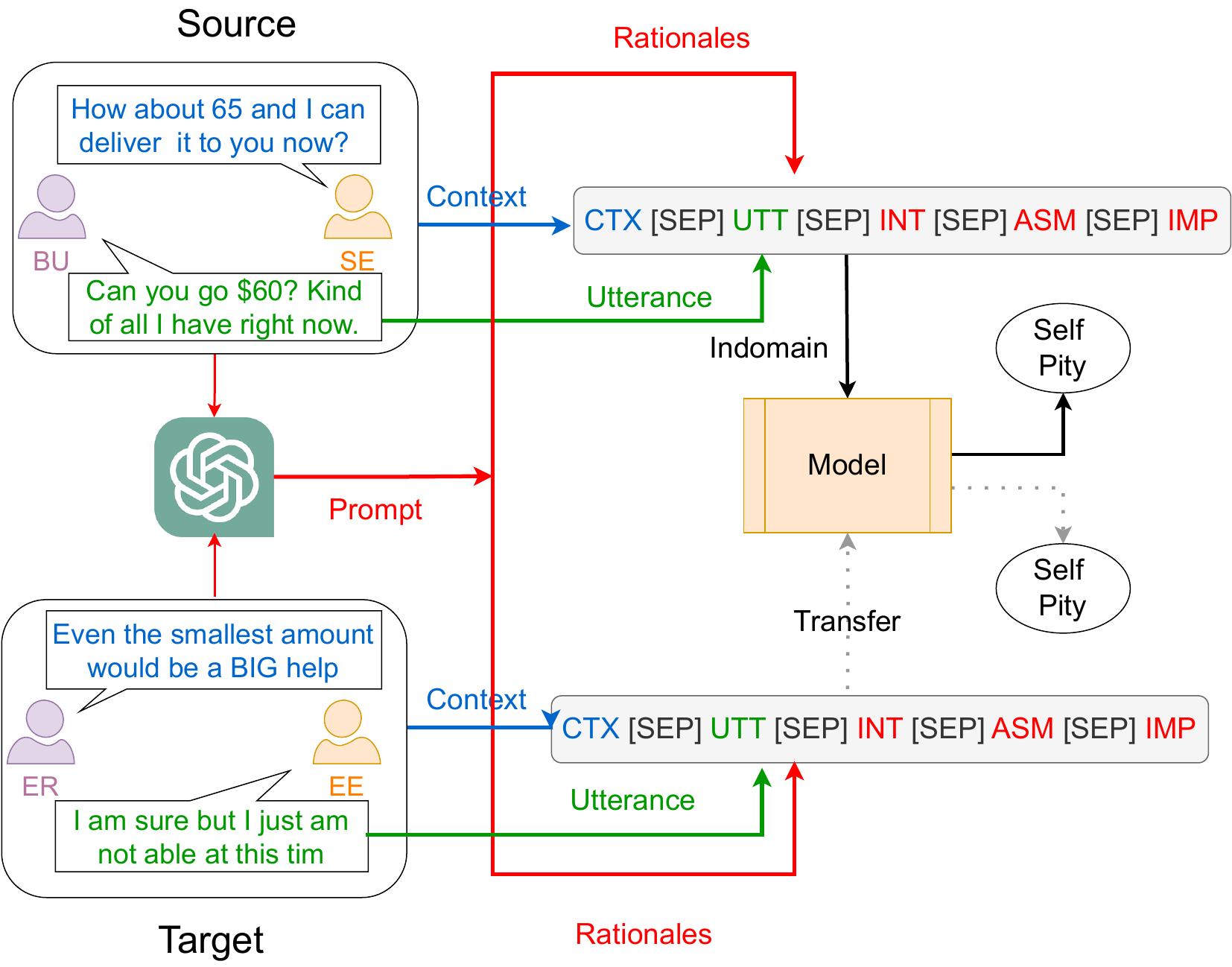}
    \caption{Here we illustrate the process of transfer from the source to target. The model is first fine-tuned on the source dialogues, which comprises the current utterance, the previous dialogue context, and the rationales (INT, ASM, and IMP for intentions, assumptions, and implicit information respectively). This fine-tuned model can then be used off-the-shelf for predictions on the target (zero-shot) or further fine-tuned in a few-shot setting.}
    \label{fig:methodology}
\end{figure*}

\subsection{Settings: In-domain and Transfer}
We carry out our experiments in two key settings, namely (i) in-domain (or ID) where the model is evaluated on unseen instances from the same domain or dataset as during training, and (ii) transfer (or TF) where a model that is first finetuned on a domain (say CB) is subsequently used for inference/training on another domain (say P4G). 

For both ID and TF scenarios, we simply pass to the model, the concatenated text comprising the past conversational context (whenever applicable), the current utterance, and one or more generated rationales corresponding to the utterance each separated by a [SEP] token. Our baseline is thus simply the text without the generated rationales. For examples, where the generated rationales are invalid, we treat them similar to our baseline.

Additionally, we replicate the experiments for both ID and TF for different N-way, k-shot cases, where k $\in$ 5, 10, 20, 50, and 100. This enables us to diagnose the impact of adding rationales while controlling for data sparsity. 

\subsection{Models and Metrics}

We explore both fine-tuning and few-shot prompting, with the latter being used for inference.

\noindent \textbf{Fine-tuning}: We fine-tune three distinct language model families ubiquitous for most NLP applications like \citet{feta}.

 \textbf{(i) Encoder only:} We use the base-uncased-version of BERT \cite{devlin-etal-2019-bert} 

\textbf{(ii) Decoder only:} We employ the base-version of GPT2 \cite{gpt2}. 

\textbf{(iii) Encoder-Decoder:} We utilize the base-version of T5 \cite{t5}.

\noindent\textbf{Few-shot prompting:} We also explore the ability of LLMs, both proprietary and open-source, in a few-shot learning setting. We experiment with GPT-3.5-turbo-16k and the Llama-2-13b-chat-hf  \cite{touvron2023llama}. We carry out inference in 0-shot and 5-shot setting for LLama-2. We consider only 0-shot for ChatGPT, due to budget restrictions. For 5-shot we randomly sample five positive and five negative instances for a given category from the training split and append them after the task description and instruction. The few-shot prompting framework appears in Table \ref{table:prompt} in the Appendix.

\noindent\textbf{Metrics:}  For all settings, we evaluate task performance in terms of the macro-averaged F1 score to account for the uneven distribution of labels for the dataset. We reproduce our experiments across three seeds and report the mean $\pm$ std deviation.

\noindent\textbf{Statistical Analysis:} We perform statistical significance using the paired bootstrapped test of \citet{berg-kirkpatrick-etal-2012-empirical} to compare model performance in presence of rationales against the corresponding baseline (absence of any rationale) as stated in \citet{dror-etal-2018-hitchhikers}. 

\section{Results}


\begin{table*}[h]
\centering
\caption{Performance of the base-variants of models (BERT, GPT2, and T5) on all 4 datasets in an in-domain setting for the entire dataset over three seeds. The rationales (RAT) 
correspond to intention (INT), assumption (ASM), implicit information (IMP), and the combination of all 3 (ALL) while the absence of any rationale is denoted by -. The best performance for each model category and dataset is denoted in bold, while * signifies the model performs significantly better than the baseline (only the utterance or -).  }
\resizebox{\linewidth}{!}{
\begin{tabular}{l| c c c | c c c |c c c |c c c}
\toprule

 & \multicolumn{3}{c}{CB} & \multicolumn{3}{c}{P4G} & \multicolumn{3}{c}{friends} & \multicolumn{3}{c}{IEMOCAP} \\
 \midrule
 RAT & BERT & GPT2 & T5 & BERT & GPT2 & T5 &  BERT& GPT2 & T5 & BERT & GPT2 & T5 \\ \midrule
- & 66.7±3.6 & 60.0±0.9 & 70.8±1.8 & 50.6±2.5 & 35.7±4.4 & 48.8±0.9 & 40.9±0.9 & 26.5±0.8 & 39.8±3.4 & 40.7±1.5 & 35.3±2.4 & 42.8±1.7 \\
INT & \textbf{68.4±1.7} & 65.6±2.0* & 70.6±2.8 & 53.0±1.6 & 45.7±1.6* & 51.2±1.4 & 45.3±0.8* & 44.5±1.0* & \textbf{44.8±2.6} & \textbf{42.6±1.3} & \textbf{42.5±2.4*} & \textbf{45.0±0.7*} \\
ASM & 66.6±0.7 & 65.3±1.3* & 69.0±1.8 & 49.4±8.1 & 47.7±2.4* & 51.1±0.8 & 44.6±0.1* & 43.4±1.2* & 39.8±0.6 & 41.0±1.8 & 39.3±3.2* & 43.1±0.6 \\
IMP & 66.9±0.3 & 64.9±1.6* & 69.1±2.6 & 52.3±1.7 & 50.1±2.6* & 51.7±3.0* & 44.7±1.7* & 43.3±1.9* & 44.1±3.3 & 42.0±1.2 & 39.9±0.9* & 42.0±0.8 \\
ALL & 67.0±0.7 & \textbf{66.0±1.5*} & \textbf{72.2±0.5} & \textbf{53.2±1.4} & \textbf{50.1±1.4*} & \textbf{53.4±2.7*} & \textbf{46.2±1.3*} & \textbf{45.5±0.8*} & 43.8±3.1* & 40.4±1.0 & 39.7±1.8* & 44.2±1.2 \\
\bottomrule
\end{tabular}}
\label{tab: ID-results}

\end{table*}

\noindent{\textbf{[RQ1:] What is the impact of rationales on task performance for the in-domain  (ID) setting?}}

We present the results of incorporating rationales on all four datasets for the supervised fine-tuned models in an  in-domain setting in Table \ref{tab: ID-results}. We observe that adding rationales improves model performance over that achieved by the baseline that uses only the utterance. The best F1 score is observed with the combination of all three rationales (ALL) followed by intention (INT). 

A more nuanced view reveals that T5 achieves the best task performance followed by BERT and then GPT2. However, we notice a disparate impact of adding rationales on different language model families. GPT2 show significant and consistent improvements across all datasets in presence of any rationale.  T5 also benefits largely from rationales where the best ID performance is significant for 3 datasets. In contrast, BERT shows significant performance over the baseline only on the ``Friends'' dataset. We posit that this could be due to higher quality of rationales generated for the ``Friends''.

\begin{figure*}
\centering
    \includegraphics[width=\linewidth]{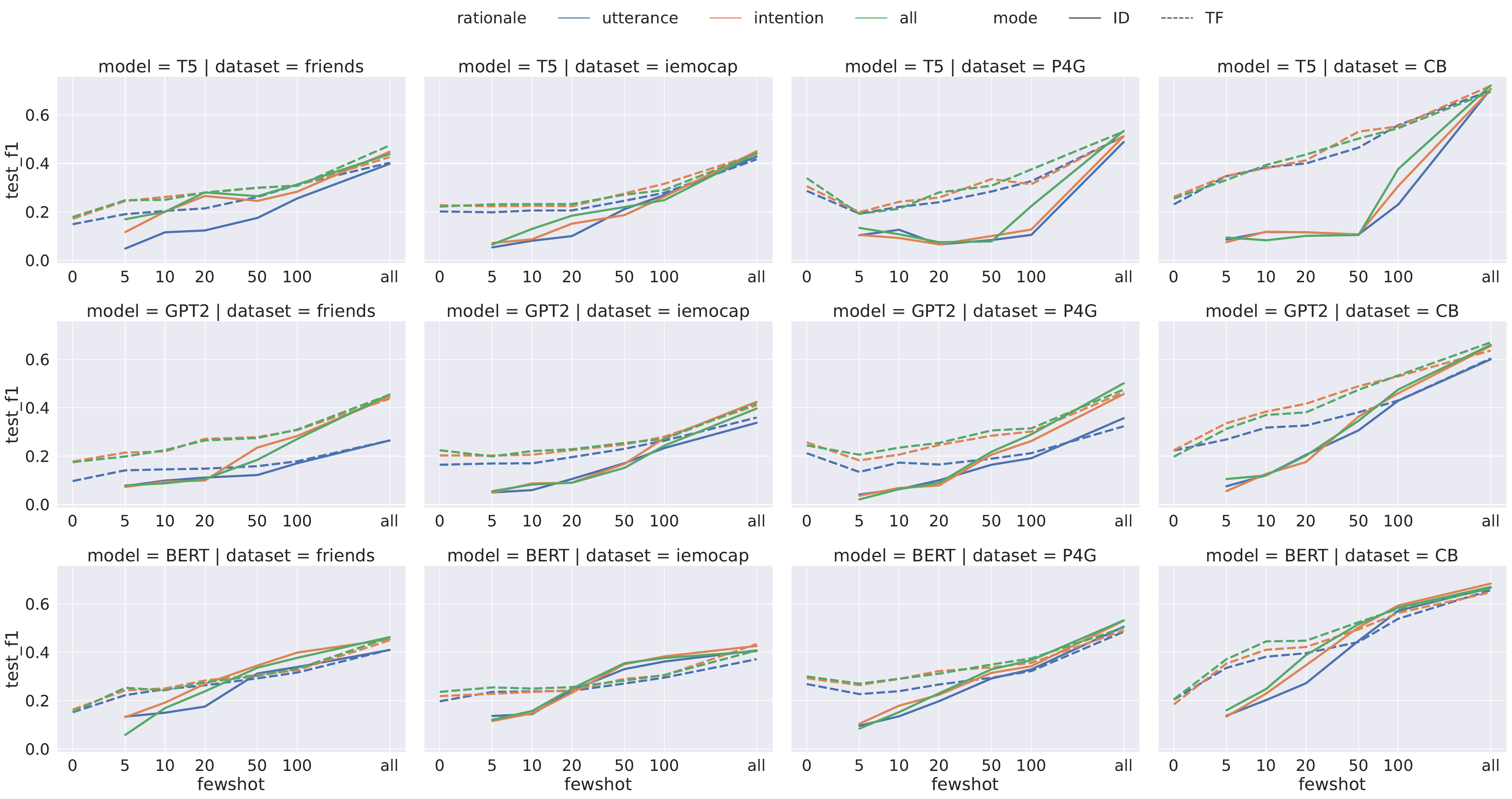}
    \caption{Performance of the base-variants of models (BERT, GPT2, and T5) on the four datasets for different few-shot examples. The solid and dashed lines correspond  to the indomain (ID) and transfer (TF) case respectively. }
    \vspace{-0.3cm}
    \label{fig:TL-experiments}
\end{figure*}

\begin{table*}[h]
\centering
\caption{Task performance in a few-shot prompting setting; 0-shot for GPT-3.5-turbo-16k (GPT-3.5), and both 0-shot and 5-shot for the 13B variant of LLama2-chat model (LLama2-0 and LLama2-5 respectively) . The rationales (RAT) correspond to intention (INT), assumption (ASM), implicit information (IMP), and all 3 (ALL) while the absence of any rationale or the baseline is denoted by -. The best performance for each model is highlighted in bold.}
\resizebox{\linewidth}{!}{
\begin{tabular}{l| c c c | c c c |c c c |c c c}
\toprule
 & \multicolumn{3}{c}{CB} & \multicolumn{3}{c}{P4G} & \multicolumn{3}{c}{Friends} & \multicolumn{3}{c}{IEMOCAP} \\
 \midrule
RAT & GPT-3.5 & LLama2-0 & LLama2-5 & GPT-3.5 & LLama2-0 & LLama2-5 & GPT-3.5 & LLama2-0 & LLama2-5 & GPT-3.5 & LLama2-0 & LLama2-5 \\ \midrule
- & 29.6 & 18.9 & 18.7 & 39.3 & 1.1 & 20.3 & 33.0 & 18.4 & 20.2 & 23.8 & 16.0 & 22.4 \\
INT & 31.3 & 14.4 & 21.5 & 40.2 & 1.5 & 19.1 & 37.7 & \textbf{24.3} & 24.9 & 26.5 & \textbf{25.6} & 23.6 \\
ASM & 31.2 & 16.2 & 21.4 & 39.6 & 5.8 & 19.7 & 38.8 & 20.4 & 23.6 & 26.2 & 25.2 & 22.5 \\
IMP & 31.9 & 18.8 & \textbf{23.2} & 39.7 & 6.6 & \textbf{27.7} & 39.5 & 22.2 & 23.2 & 26.5 & 24.5 & \textbf{24.7 } \\
ALL & \textbf{32.4} & \textbf{19.2} & 19.2 & \textbf{41.2} & \textbf{9.9} & 20.9 & \textbf{39.9} & 23.3 & \textbf{32.5} & \textbf{27.0} & 24.8 & 23.1 \\
\bottomrule
\end{tabular}}
\label{tab:ICL-experiments}
\end{table*}

\noindent{\textbf{[RQ2:] How does adding rationales influence few-shot task performance?}}

We present our results for incorporating rationales on task performance for both in-domain (ID) and transfer (TF) for different k-shot cases in Figure \ref{fig:TL-experiments}. We restrict our findings to rationales corresponding to intention (INT) and combination of all three (ALL) because they had the highest performance in Table \ref{tab: ID-results}. Our complete set of results are relegated to Figure \ref{fig:all-info-lineplot} in the Appendix.

\noindent \textbf{Impact of transfer:} One key finding is that the TF performance is consistently higher than in ID (dashed lines score better than the corresponding solid lines) possibly because the model is already trained on the entire source dataset. This is more pronounced in the low data regimes for k-shot corresponding to 5, 10, 20, and 50. and is consistent across all pairs of model and dataset combinations. However, the gain diminishes by the time the model is fine-tuned on the entire dataset (denoted by 'all'). 

Moreover, adding rationales is better realized for  TF than ID;  73.8\% of all TF experiments with the rationale ALL had a significantly higher performance over the baseline, while only 5.9\% experiments were statistically worse than the baseline. Compare this with 57.0\% and 19.4\% for ID. 

\noindent \textbf{Impact of rationales:} Another key finding is the disparate impact of rationales on the task choice. ERC benefits more than RES from adding rationales. For TF, 85.7\% and 60.7\% of cases that include the rationales are significantly better for ERC and RES respectively; the corresponding proportion in the ID setting is 55.5\% and 51.4\% respectively. We posit that since the semantic meaning of emotions remains consistent across domains, rationales facilitate transfer better for ERC; or alternately ERC is an easier task than RES. 

This observation is echoed vividly in 0-shot transfer where we observe a significant gain 83.3\% of the times for ERC as opposed to 41.7\% for RES. Nevertheless, in a few-shot setting when the model is exposed to instances from the corresponding target domain, the gains start racking up. We emphasize that across all experiments,  rationales perform significantly worse than the baseline fewer than 10\%. Thus, from a big picture view, rationales can indeed facilitate task performance and transfer. 

\noindent\textbf{Significant Testing:} Considering our massive slew of 2,340 experiments, spanning multiple datasets, models, few-shot cases, rationales, and modes (ID/ TF) we also conduct a full-factorial analysis of the experimental suite to obtain a conservative estimate of statistical significance that incorporates the needed adjustments in the face of multiple comparisons in order to avoid type I errors \cite{gururaja2023linguistic}.  For each task, we computed an ANCOVA model with task f1 as the dependent variable, with model (BERT, T5, and GPT2), mode (ID vs TF), rationale (none, INT, ASM, IMP, and ALL) and target domain as independent variables, and few-shot setting nested within mode as a covariate.  We also included all 2-way and 3-way interactions between independent variables in the model.  

For RES, all independent variables and the covariate were significant, but not the interactions between independent variables. Moreover, performance on CB was consistently higher than P4G, with BERT being the best model.  ID was consistently worse than TF.  ALL was the best rationale setting, with ASM being the only rationale  that was significantly worse than ALL.  Including no rationale was significantly worse than all other rationale settings except for ASM. 

The story is a little more complicated for the ERC task.  We have all the same main effects except dataset -- for this task, they are not different from one another.  ALL and INT were equally good, and both better than IMP and ASM.  All of these were significantly better than including no rationale.  There was an interaction between model and these rationales such that the ordering of preferred rationale setting was relatively consistent across different models, but which contrasts were significant varied (note the Tables in the Appendix where different models achieve the best score with different rationales).  Nevertheless, including rationales was always better than not including rationales at all, and INT was consistently ranked high. In a nutshell, the rationale INT had the highest impact on model performance.

\noindent{\textbf{[RQ3:] How does adding rationales affect few-shot prompting performance for LLMs?}}

We present our results of using rationales for few-shot prompting in LLMs in Table \ref{tab:ICL-experiments}. We observe similar trends to the supervised learning set-up wherein the inclusion of rationales improves task performance. 
Once again, the combination of rationales (ALL) achieves the highest F1 score, while both INT and IMP take a close second. Unsurprisingly, we see the best performance for GPT-3.5 in 0-shot followed by LLama2-13B in a 5-shot setting. Nevertheless, the few-shot prompting results are significantly worse than the fine-tuned supervised models, with results on CB and IEMOCAP being matched by our smaller models at k=5 and k=50 respectively.







\section{Qualitative Analysis}

Having demonstrated the efficacy of rationales to facilitate understanding of social meaning in dialogue, we do a deep dive on their utility, namely where do rationales help and why. 

We investigate the impact rationales have on individual task labels or strategies in ID. For each dataset, we consider the combination of model and rationale pair with the highest ID performance in Table \ref{tab: ID-results} and compare their predictions against the baseline (the corresponding model with only UTT). Immediately, we observe that rationales help to shift or re-distribute the prediction probability mass from the majority (``neutral'' for ERC and ``Not a resistance strategy or NAS'' for RES) to others. 

We highlight examples where adding rationales were consistently better in  Table \ref{table:utterance_analysis_better} and cases where their presence consistently degrades performance in Table \ref{table:utterance_analysis_worse}. In the following analysis we refer to instances in these Tables in the Appendix.

\noindent\textbf{Rationales perform better for ERC:} Notably, for ERC, adding rationales is better at identifying the emotions ``surprise'' and ``anger''. This improved performance can be largely attributed to the fact that the elicited rationales, particularly the intentions (INT), make apparent the emotional state. For instance, the INT rationale interprets the exclamation mark ``!'' in the utterance for the Friends dataset as an expression of excitement or surprise, and thus corresponds with the actual label (surprise). Likewise, for the utterance ``Thanks'' from IEMOCAP is characterized in the rationales as reflecting gratitude or acknowledgment of support and condolences, contributing to an overall sentiment of ``sadness'' in response to a bereavement consolation.

\noindent\textbf{Rationales perform worse for ERC:} The cases where the model mispredicts can be linked to the specific language usage. For example, the utterance in friends ``What the hell happened on that beach?!'' is erroneously interpreted as anger possibly due to ``what the hell.'' Likewise, for the utterance ``I'm just worried,'' in IEMOCAP, the rationales express a sense of anxiety or uncertainty from ``worried'' misleading the prediction as ``other'' than ``sadness.''

\noindent\textbf{Rationales perform better for RES:} For RES, the integration of rationales notably enhances performance for ``Counter Argumentation'' and ``Hesitance.'' E.g., in the CB dataset, for the utterance ``but how about 180 since I'm the one picking it up and with its one handle missing?'', the rationale accurately identifies the buyer's intention to propose a reduced price due to the item's missing handle, and thus aligns with Counter Argumentation. Furthermore, for P4G, ``when finished with this task I will be sure to check the website,'' the rationales portray the speaker's implied conditional interest, indicating Hesitance as the action is deferred until task completion. 

\noindent\textbf{Rationales perform worse for RES:} Conversely, the model's performance for the ``Source Derogation'' strategy is less effective. A typical example is ``perhaps a link to an organization or other agency that rates major charities would be more helpful'' for P4G. Here, the rationales inaccurately interpret the statement as a mere suggestion for a more efficient information source, and fail to detect the speaker's skepticism about the organization's credibility. We posit that this misprediction is linked to LLM's tendency to generate responses with a positive connotation, leading to a misinterpretation of critical tones as constructive suggestions. This results in erroneous labeling as ``Information Inquiry'' indicating a request for additional information, or ``Counter Argumentation,'' which suggests an alternative factual proposition.

While we note that overall rationales facilitate transfer, the gains observed are not symmetric. Specifically, we observe higher gains for the less frequent classes in the target dataset, such as the emotion ``fear'' on Friends and ``Source Derogation'' and ``Self Pity'' classes on the P4G dataset.


\section{Conclusion and Future Work}
We present a generalizable framework that leverages machine-generated rationales from LLMs to deduce the underlying social meaning embedded in conversations. We observe that augmenting pretrained models with the generated rationales significantly improves performance over the baseline across multiple datasets for both the tasks of emotion recognition and detecting resisting strategies. The gains are pronounced during cross-domain transfer across both zero-shot and few-shot settings thereby highlighting the generalizability of our approach. While our current work place emphasis on domain adaptation, we believe the proposed approach is generalizable to new social meaning detection tasks (persuasion, empathy, argumentation) which we defer for future work. Furthermore, as opposed to leveraging an LLM, we intend to deploy or instruct-tune smaller models that can generate these rationales \cite{what-makes-it-ok-2023, zhou-etal-2023-cobra}.

\section*{Limitations}

Some of the main limitations of our work include:

(i) Reliance on closed-source or proprietary LLMs to generate rationales. Consequently we are not able to assure the reproducibility of generating the rationales or whether the service will be discontinued. We do however, release the entire dataset of rationales for public use for reproducibility. 

(ii) We note that our proposed framework of generating rationales for fine-tuning a smaller model can be deemed more expensive than approaches that just prompts the LLM for an answer while generating these rationales during inference \cite{wei2022chain}. However, one of our contributions was to demonstrate that our approach is indeed possible. In a future work, we intend to use our created dataset, to instruction tune a smaller LM, like Flan-T5 \cite{flan-t5} to generate these rationales in-house. Prior work has demonstrated the reliability of this approach \cite{what-makes-it-ok-2023, zhou-etal-2023-cobra}, and we intend to follow up in a future work for other social meaning detection tasks like persuasion, negotiation, empathy amongst others. 

(iii) Recent studies, including \cite{zhou-etal-2022-prompt, sclar2023quantifying, leidinger2023language}, highlight prompt sensitivity and the influence of prompt choice on downstream tasks. Our manual evaluation of 80 GPT-3.5-generated rationales, using our selected prompts, indicates they are of sufficient high quality. Potential prompt optimization avenues may exist for further enhancing rationale quality, but we defer exploration to future work.

(iv) Our choice to limit investigation to two datasets and three models is a deliberate  one aimed at managing computational resources. Even within this constrained framework, we conduct 2340 experiments, highlighting the substantial computational demands of our analysis.

(v) We employ GPT-3.5-generated rationales in our study. However, we remain uncertain about their status as the ideal rationales for this purpose, or which kinds of rationales are the most effective towards this particular task.


\section*{Ethical Concerns}

Our research relies on the responses generated by LLMs which are known to exhibit hidden biases in their representations. While during our experiments, we encountered no potential biases in terms of offensive language or stereotypes in the generated response for our controlled setting of social meaning detection, we implore practitioners and other researchers to conduct thorough analysis before adopting our particular prompting approach for the respective use-case. We also recognize the limitations of LLM in interpreting social meanings and clarify that our conclusions, based on probabilistic model outputs, do not construe absolute facts. Moreover, we stress that the application of LLM rationales, while beneficial within our controlled research environment for understanding human intent in utterances, should not be extended uncritically beyond these confines. The use of LLM rationales in broader contexts, especially as substitutes for human judgment and rationale, is not advocated.

\section*{Acknowledgments}

This work was funded in part by NSF grants DUE 2100401 and ITEST  2241669.
We also thank the reviewers and program chairs for their helpful comments.

\bibliography{anthology,custom}
\bibliographystyle{acl_natbib}

\appendix

\clearpage
\newpage

\section{Dataset Statistics}
Figure \ref{fig: datasets-label-distribution} provides a distribution of labels for the two tasks of ERC and RES across the respective two datastes.  Furthermore, Table \ref{tab:res-framework} and Table \ref{tab:erc-framework} provide additional 
insight into the definition of the categories/strategies for the corresponding datasets, as well as representative examples of the same. 

Table \ref{tab: datasets-rationales-stats} also presents statistics of the datasets and the corresponding rationales. Each dialog is broken into multiple datapoints, one for each turn in it. The average number of turns per dialogue and the number of words per turn are reported, with IEMOCAP seen to have significantly longer dialogues compared to the rest. The number of rationales generated for the dataset are reported -- For P4G and CB, we encounter parsing issues with GPT-3.5's generated rationales for some instances, which are ignored during training. The average number of words per generated intention/assumption/implicit information is higher for the emotion datasets compared to the resisting strategies ones, which may have been influenced by the choice of the one-shot example in the prompt. The generated implicit information is found to be longer than intention and assumption, and assumption is found to be longer than intention, across all datasets.

\begin{table}[h]
\centering
\caption{Fraction of times ChatGPT-3.5-turbo-16k was chosen over LLama-2-13B-chat based on the quality of the generated rationales.}
\begin{tabular}{l c c c c}
\toprule
 & CB & P4G & Iemocap & friends \\
 \midrule
S1 & 15 & 16 & 12 & 16 \\
S2 & 13 & 15 & 14 & 19 \\
S3 & 13 & 11 & 12 & 12 \\
Overall & 15 & 16 & 12 & 17 \\
\bottomrule
\end{tabular}

\label{tab: chatgpt-vs-llama-ratgen}
\end{table}

\begin{table}[h]
\centering
\caption{Hyperparameters used for fine-tuning the models (BERT, T5, and GPT2) across all datasets.}
\begin{tabular}{l|r}
\toprule
\textbf{Hyperparameter} & \textbf{Value}\\
\midrule
Max sequence length&512\\
Learning rate&$2e^{-5}$\\
Batch size&16\\
Num. epochs&15\\
Optimizer & Adam\\
Patience & 5 \\
\bottomrule
\end{tabular}
\label{appendix: hyperparams-finetuning}
\end{table}

\begin{table*}[]
\centering
\begin{tabular}{l|ll|ll}
\toprule
                           & \multicolumn{2}{c}{\textbf{ERC}} & \multicolumn{2}{c}{\textbf{Res}} \\ \midrule
                           & \textbf{Friends}    & \textbf{IEMOCAP}    & \textbf{P4G}        & \textbf{CB}         \\ \midrule
Dialogues                  & 1000       & 151        & 473        & 713        \\ \hline
Total datapoints           & 14503      & 10039      & 11260      & 8511       \\ \hline
Labels                     & 8          & 8          & 8          & 8          \\ \hline
Avg. Turns/Dialogue             & 14.50      & 66.49      & 36.05      & 11.94      \\ \hline
Avg. Words/Turn                 & 7.83       & 11.57      & 9.22       & 12.38      \\ \midrule
Rationales Generated       & 97.8\%      & 94.78\%      & 97.90\%      & 86.38\%       \\ \hline
Avg. Words/Intention            & 32.56      & 24.47      & 15.00      & 14.07      \\ \hline
Avg. Words/Assumption           & 39.06      & 31.79      & 17.46      & 15.10      \\ \hline
Avg. Words/Implicit Information & 50.04      & 44.29      & 19.41      & 16.55      \\ \bottomrule
\end{tabular}
\caption{We present here the statistics of the datasets used and the rationales generated.}
\label{tab: datasets-rationales-stats}
\end{table*}


\begin{table*}[]
\small
\caption{Framework describing the resisting strategies for persuasion (P4G) and negotiation (CB) datasets, as specified in \citet{dutt-etal-2021-resper}. Examples of each strategy are italicised. The examples for each of P4G and CB were borrowed from the original datasets of the same name from \citet{wang-etal-2019-persuasion} and \citet{he-etal-2018-decoupling} respectively.  }
\centering

\begin{tabular}{p{0.16\textwidth} p{0.37\textwidth}p{0.38\textwidth}}
\toprule
Resisting Strategy & Persuasion (P4G) & Negotiation (CB)\\ \midrule

Source Derogation & Attacks/doubts the organisation's credibility. & Attacks the other party or questions the item.\\ 

& \textit{My money probably won't go to the right place}& \textit{Was it new denim, or were they someone's funky old worn out jeans?} \\ \hline

Counter Argument & Argues that the responsibility of donation is not on them or refutes a previous statement.  & Provides a non-personal argument/factual response to refute a previous claim or to justify a new claim. \\ 

&\textit{There are other people who are richer} & \textit{It may be old, but it runs great. Has lower mileage and a clean title.}\\ \hline

Personal Choice & Attempts to saves face by asserting  their personal preference such as their choice of charity and their choice of donation. & Provides a personal reason for disagreeing with the current situation or chooses to agree with the situation provided some specific condition is met.\\
& \textit{I prefer to volunteer my time} & \textit{I will take it for \$300 if you throw in that printer too.}\\  \hline

Information Inquiry & Ask for factual information about the organisation for clarification or as an attempt to stall. & Requests for clarification or asks additional information about the item or situation. \\
& \textit{What percentage of the money goes to the children?} & \textit{Can you still fit it in your pocket with the case on?}\\ \hline

Self Pity & Provides a self-centred reason for not being able/willing to donate at the moment. & Provides a  reason (meant to elicit sympathy) for disagreeing with the current terms.\\
& \textit{I have my own children} & \textit{\$130 please I only have \$130 in my budget this month.}\\ \hline

Hesitance & Attempts to stall the conversation by either stating they would donate later or is currently unsure about donating.  & Stalls for time and is hesitant to commit; specifically, they seek to further the conversation and provide a chance for the other party to make a better offer. \\
&\textit{Yes, I might have to wait until my check arrives.} & 
\textit{Ok, would you be willing to take \$50 for it?}\\ \hline

Self-assertion & Explicitly refuses to donate without even providing a factual/personal reason & Asserts a new claim or refutes a previous claim with an air of finality/ confidence.\\ &\textit{Not today} & \textit{That is way too little.}\\ \bottomrule
\end{tabular}

\label{tab:res-framework}
\end{table*}

\begin{table*}[]
\small
\caption{Framework describing the emotion labels in the emotion recognition datasets (IEMOCAP and Friends) \cite{iemocap, poria-etal-2019-meld}. Examples of each label are italicised.}
\centering

\begin{tabular}{p{0.16\textwidth} p{0.37\textwidth}p{0.38\textwidth}}
\toprule
Emotion & IEMOCAP & Friends\\ \midrule

Neutral & Neutral emotion is characterized by the absence of strong feelings or emotions. & Neutral emotion is characterized by the absence of strong feelings or emotions.\\ 

& \textit{I'll go to basketball games.}& \textit{Yeah, apparently they're turning it into some kinda coffee place.} \\ \hline

Joy & Joy is a feeling of extreme gladness, delight, or exultation of the spirit arising from a sense of well-being or satisfaction.  & Joy is a feeling of extreme gladness, delight, or exultation of the spirit arising from a sense of well-being or satisfaction. \\ 

&\textit{I don't know it seemed like a pretty good spot to me. Look at the moon - view the moon view I got from here.} & \textit{I'm so proud of you.}\\ \hline

Sadness & Sadness is an emotional state of unhappiness, ranging in intensity from mild to extreme and usually aroused by the loss of something that is highly valued & Sadness is an emotional state of unhappiness, ranging in intensity from mild to extreme and usually aroused by the loss of something that is highly valued\\
& \textit{Augie, I'm sorry.} & \textit{Uh, well... Joey and I broke up.}\\  \hline

Surprise & Surprise is an emotion typically resulting from the violation of an expectation or the detection of novelty in the environment. & Surprsie is an emotion typically resulting from the violation of an expectation or the detection of novelty in the environment. \\
& \textit{Shut up. No- in Vegas?} & \textit{Oh my God, wh-what happened?}\\ \hline

Fear & Fear is a basic, intense emotion aroused by the detection of imminent threat, involving an immediate alarm reaction that mobilizes the organism by triggering a set of physiological changes. & Fear is a basic, intense emotion aroused by the detection of imminent threat, involving an immediate alarm reaction that mobilizes the organism by triggering a set of physiological changes.\\
& \textit{Good God.} & \textit{Oh boy, I just can't watch. It's too scary!}\\ \hline

Disgust & Disgust is characterized by strong aversion to something deemed revolting, or toward a person or behavior deemed morally repugnant. & Disgust is characterized by strong aversion to something deemed revolting, or toward a person or behavior deemed morally repugnant. \\
&\textit{It was a terrible thing. I hated it.} & 
\textit{Ew! What is that? Something exploded!}\\ \hline

Other & An emotion or feeling which does not include anger, surprise, sadness, joy, fear, or disgust. & An emotion or feeling which does not include anger, surprise, sadness, joy, fear, or disgust.\\ &\textit{How long did that row last?} & \textit{Oh well, okay, good luck.}\\ \bottomrule
\end{tabular}
\label{tab:erc-framework}
\end{table*}

\section{Qualitative Analysis of Rationales}
\label{sec: annotation analysis}
We present qualitative analysis of the responses generated by LLMs (GPT-3.5-turbo-16k and LLama2-13B-chat-hf) here with Table \ref{tab: chatgpt-vs-llama-ratgen} highlighting the fraction of times the annotators preferred the quality of response generations of  ChatGPT to LLama2. Table \ref{tab: rat-quality} highlights the average score of two annotators on the qaulity of responses generated across different datasets in terms of grammaticality, relevance, and factuality. 

Grammaticality is defined as how well formed, fluent, and grammatical the response is. It achieves a high score due to the sufficient prowess of contemporary LLMs on text generation. 

Relevance indicates  whether the rationale generated actually answers the prompt query, i.e. the generated rationale aligns well with a human's view of  the speaker's intention, assumption, and implicit information about the conversation.

Factuality indicates whether the rationale generated is consistent with the dialogue history; i.e. it does not hallucinate additional information or talk about cases which are not present in the text. 

We also provide examples of the actual prompt framework for the ERC and RES in Table \ref{table:erc-prompt} and \ref{table:res-prompt} respectively. 

\begin{table*}[]
\centering
\caption{Below is an example of our prompt for the task of emotion recognition in conversations (ERC).}
\small
\resizebox{\textwidth}{!}{
\begin{tabular}{p{0.2\linewidth}|p{0.7\linewidth}}
\toprule
 Part 1: High level description of the objective & Analyze this dialogue, focusing on any underlying assumptions and implicit information. \\
\midrule
Part 2: Instructions & For the final utterance, provide a comprehensive and concise explanation for: a) Speaker's Intention, b) Assumptions about the conversation, and c) Implicit Information \\
\midrule
Part 3: Output Template &Please format your response as follows: \\
&Speaker's Intention in the final utterance: <your response> \\
&Assumptions about the conversation in the final utterance: <your response> \\
&Implicit Information in the final utterance: <your response> \\
\midrule
Part 4: Examples for ICL&Dialogue history:\\
&The Interviewer: You must've had your hands full.\\
&Chandler: That I did. That I did.\\
&The Interviewer: So let's talk a little bit about your duties. \\
&Chandler: My duties?  All right.\\&\\
&Final utterance:\\
&The Interviewer: Now you'll be heading a whole division, so you'll have a lot of duties.\\
&\\
&Speaker's Intention in the Final Utterance: The speaker's intention in the final utterance is to inform Chandler about his upcoming role and the responsibilities associated with it. The speaker is preparing Chandler for a new position.\\
&\\
&Assumptions about the conversation in the Final Utterance: The assumption here is that Chandler is about to take on a leadership role within the organization, specifically heading a whole division. The speaker assumes that Chandler needs to be aware of the increased responsibilities that come with this new position.\\
&\\
&Implicit Information in the Final Utterance: The implicit information in the final utterance is that Chandler has been promoted or assigned a higher-level job within the company. Additionally, it suggests that the speaker expects Chandler to be prepared to handle the increased workload and responsibilities that come with leading a division.\\
&...\\
\bottomrule
\end{tabular}}
\label{table:erc-prompt}
\end{table*}

\begin{table*}[]
\centering
\small
\caption{Below is an example of our prompt for the task of detecting resisting strategies (RES).}
\resizebox{\textwidth}{!}{
\begin{tabular}{p{0.2\linewidth}|p{0.7\linewidth}}
\toprule
 Part 1: High level description of the objective & Analyze this dialogue, focusing on any underlying assumptions and implicit information. Ensure that you address each line individually without skipping or grouping.\\
\midrule
Part 2: Step-wise guide & For each line:\\
&1. Provide a comprehensive and concise explanation for:\\
&a)Speaker's Intention\\
&b)Assumptions about the conversation\\
&c)Implicit Information\\
&2. Continue until you have analyzed every line.\\
\midrule
Part 3: Output Template &Please format your response as follows:\\
&Speaker's Intention: <your response>\\
&Assumptions about the conversation: <your response>\\
&Implicit Information: <your response>\\
\midrule
Part 4: Examples for ICL&INPUT:\\
&...\\
&Persuadee: They are hungry and injured and also short.\\
\\
&Persuader: I'm so sorry, what a terrible thing.\\
&...\\
\\
&Output:\\
&...\\
&Speaker's Intention: The Persuadee provides additional details about their child's situation, emphasizing the child's needs.\\
&Assumptions about the conversation: The Persuadee assumes that sharing these specific details will elicit a stronger empathetic response from the Persuader.\\
&Implicit Information: The Persuadee seeks empathy and understanding from the Persuader regarding their child's dire circumstances.\\
\\
&Speaker's Intention: The Persuader expresses sympathy and acknowledges the gravity of the Persuadee's situation.\\
&Assumptions about the conversation: The Persuader assumes that offering sympathy and acknowledging the seriousness of the situation is an appropriate response.\\
&Implicit Information: The Persuader expresses compassion and understanding toward the Persuadee's plight.\\
&...\\
\bottomrule
\end{tabular}}
\label{table:res-prompt}
\end{table*}

\section{Hyperparamter Tuning}

We present the hyperparameters for our experiments in Table \ref{appendix: hyperparams-finetuning}. We carry out the experiments over 3 seeds on a A6000 GPU with early stopping with patience of 5 over the validation set for all experiments. We implement the entire experiments in Python, with help of the Pytorch library and use the pre-trained models as specified in Huggingface under the agreed upon license agreements.

Our experimental suite comprises encompasses 4 datasets in 2 settings (ID/TF) for 3 models (BERT, T5, GPT2) over 5 rationale combinations (none, INT, ASM, IMP, ALL), for 6 few-shot settings (5, 10, 20, 50, 100, and all), and re-evaluated over 3 seeds. This brings the host of experiments to 2160 experiments. There is an additional 180 cases when inferred over 0-shot TF cases, bringing the total to 2340 experiments.

The total cost of the GPT-3.5 credits during the course of our experiments totalled to approx \$250 (\$200 for generating prompts and \$50 for ICL experiments).


\begin{table*}[]
\centering
\small
\caption{Example of our prompt for the zero-shot and few-shot experiments on LLMs. We illustrate with an example from the P4G dataset.}
\resizebox{\textwidth}{!}{
\begin{tabular}{p{0.25\linewidth}|p{0.7\linewidth}}
\toprule
 Part 1: Description of strategy being referenced in the example & These examples pertains to the Source Derogation strategy. For the dataset P4G, the description of Source Derogation is as follows:\\
&Attacks/doubts the organisation's credibility.\\
\midrule
Part 2: Instructions and output format
&Given a response for a particular speaker and recent dialogue context containing the past utterances (wherever available), output 'Yes' if the utterance contains the above strategy, otherwise output 'No'. Your output should contain only 'Yes' or 'No', and no other text.\\
\midrule
Part 3: Examples for in-context learning&[CONTEXT]\\
&[Persuadee]:Would you like to donate some of your bonus payment to the charity Save the Children today? \\
&[Persuadee]:You could choose to donate any amount, from \$0 up to \$2.\\
&[Persuadee]:but i need money because i have finantial problems\\
&[Persuadee]:I do understand how that can be.\\
&[Persuadee]:But even .10 could go a long way overseas\\
&[RESPONSE]\\
&[Persuadee]:Thank you for your consideration.\\
&[INTENTION] The Persuadee expresses gratitude to the Persuader for considering their request.\\
&[ASSUMPTION] The Persuadee assumes that expressing gratitude may positively influence the Persuader's decision.\\
&[IMPLICIT INFORMATION] The Persuadee appreciates the Persuader's willingness to consider their request.\\
&[OUTPUT]\\
&No\\
&...\\
\midrule
Part 4: Test datapoint&[CONTEXT]\\
&...\\
&[RESPONSE]\\
&...\\
&[OUTPUT]\\
\bottomrule
\end{tabular}}
\label{table:prompt}
\end{table*}

\section{Additional Results}

\begin{figure*}[]
    \centering
    \includegraphics[width=1.0\linewidth]{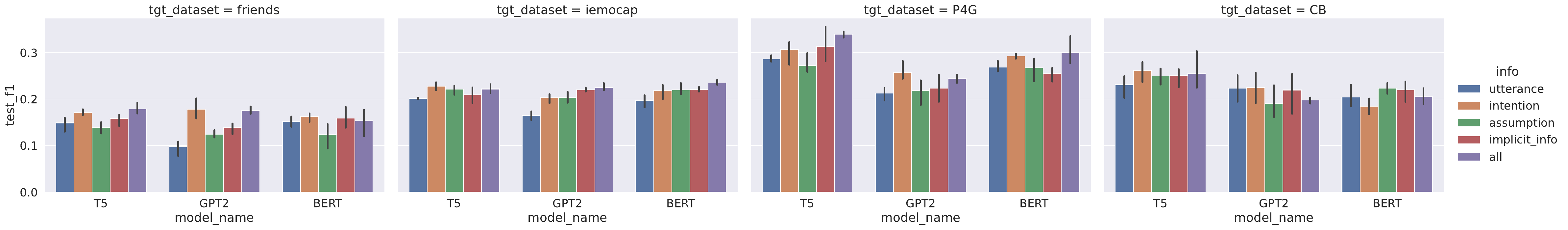}
    \caption{Performance of the base-variants of models (BERT, GPT2, and T5) on the four datasets in a zero-shot transfer setting, where models trained for the similar task on a given source domain was then applied to the new target domain (e.g. P4G $\rightarrow$ CB and CB $\rightarrow$ P4G for RES and
     friends $\rightarrow$ iemocap and iemocap $\rightarrow$ friends for ERC.)}
    \label{fig:zero-shot-perf}
\end{figure*}

\begin{figure*}[]
    \centering
    \includegraphics[width=1\linewidth]{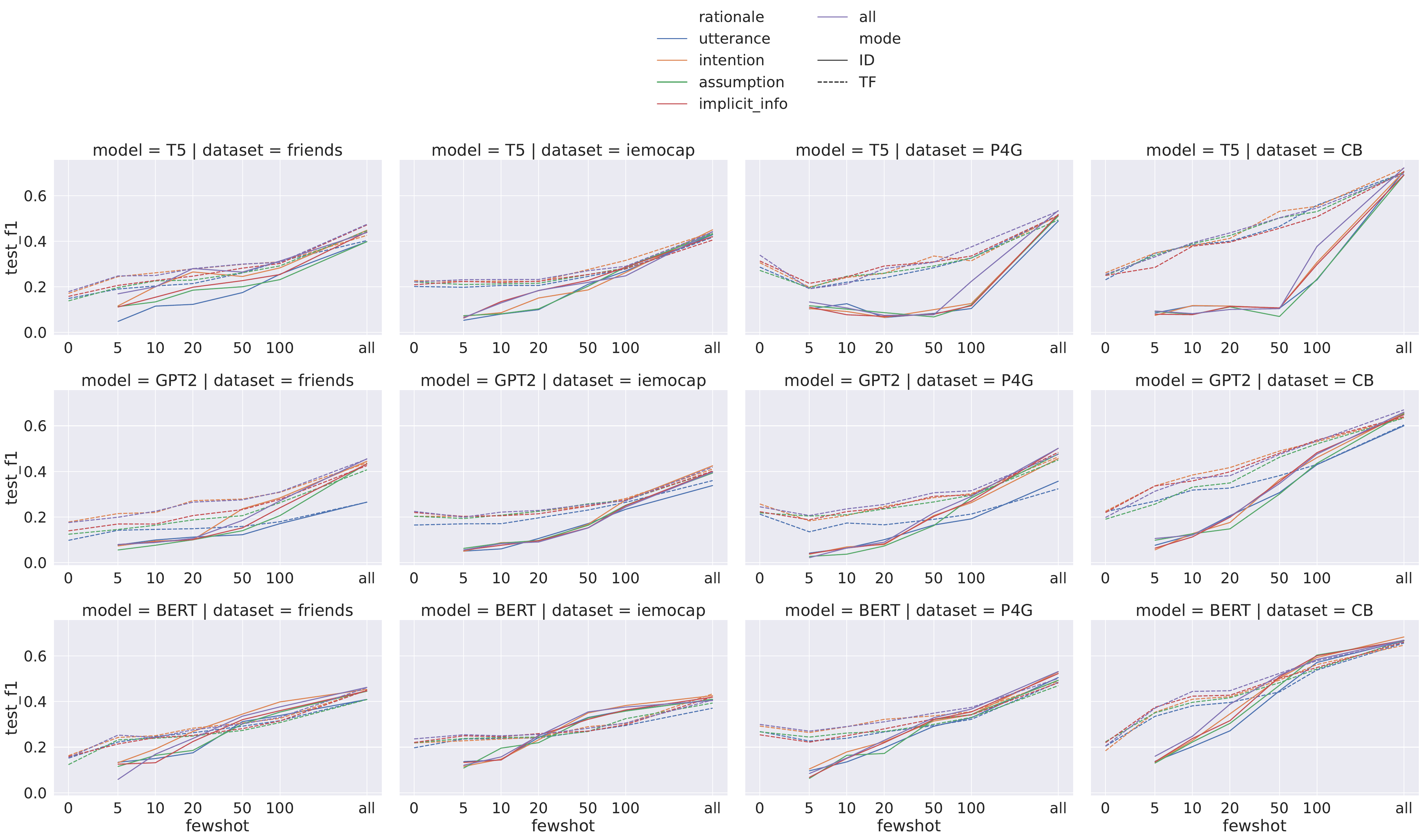}
    \caption{Performance of the base-variants of models (BERT, GPT2, and T5) on the four datasets for different few-shot examples for all rationales. The solid and dashed lines correspond  to the indomain (ID) and transfer (TF) case respectively. }
    \label{fig:all-info-lineplot}
\end{figure*}

\begin{figure*}[]
\centering
\subfloat[friends with UTT (BERT)]{\includegraphics[width=0.45\linewidth]{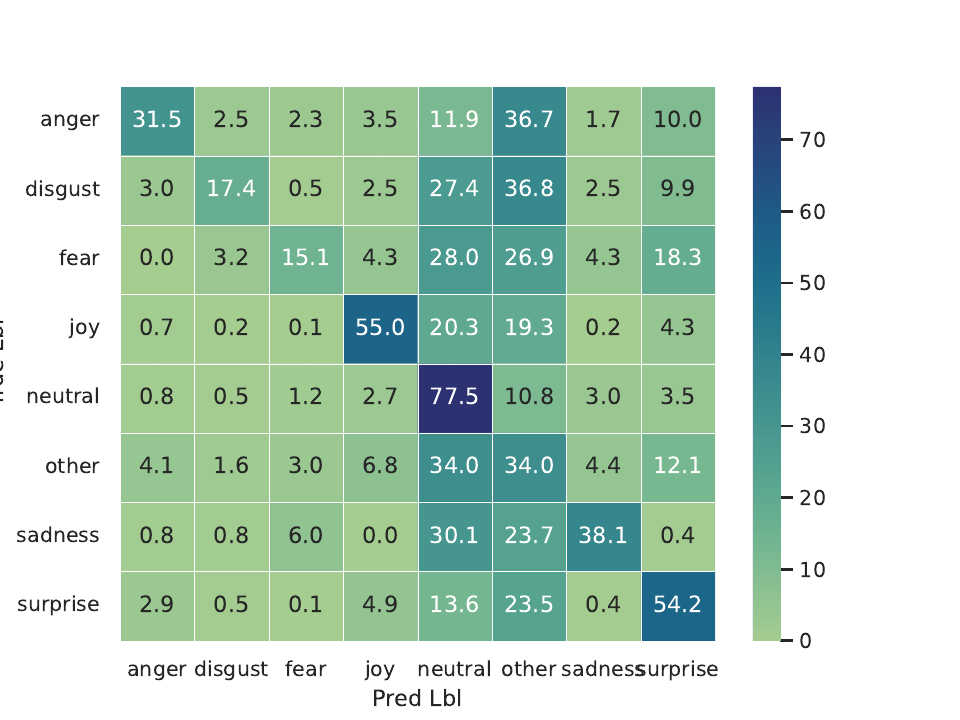}}
\subfloat[friends with ALL (BERT)]{\includegraphics[width=0.45\linewidth]{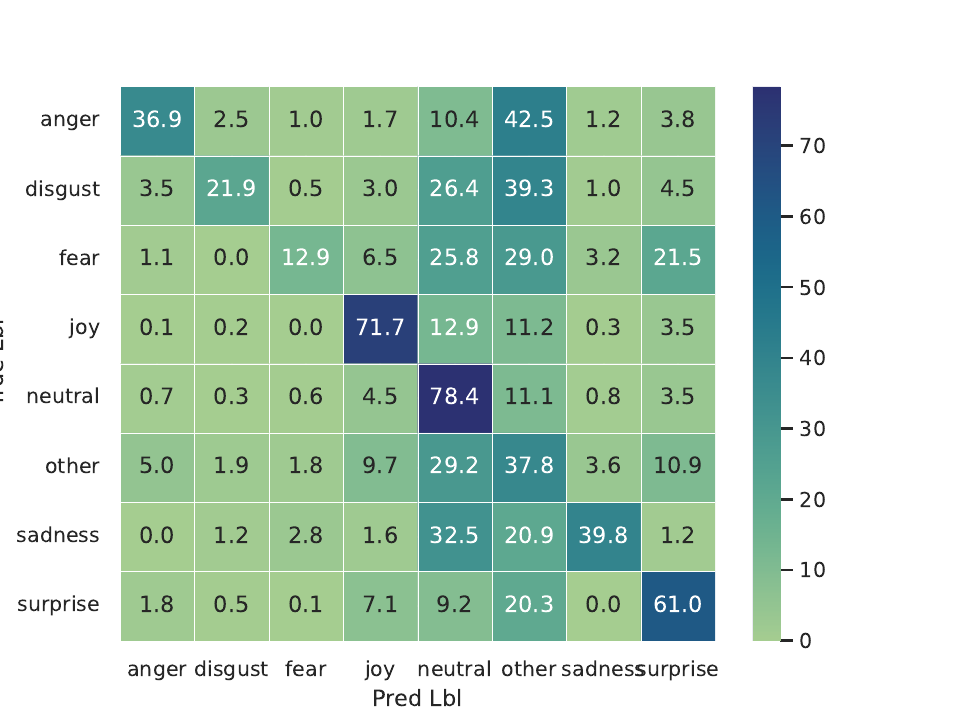}}
\vspace{-0.1cm}
\subfloat[iemocap with UTT (T5)]{\includegraphics[width=0.45\linewidth]{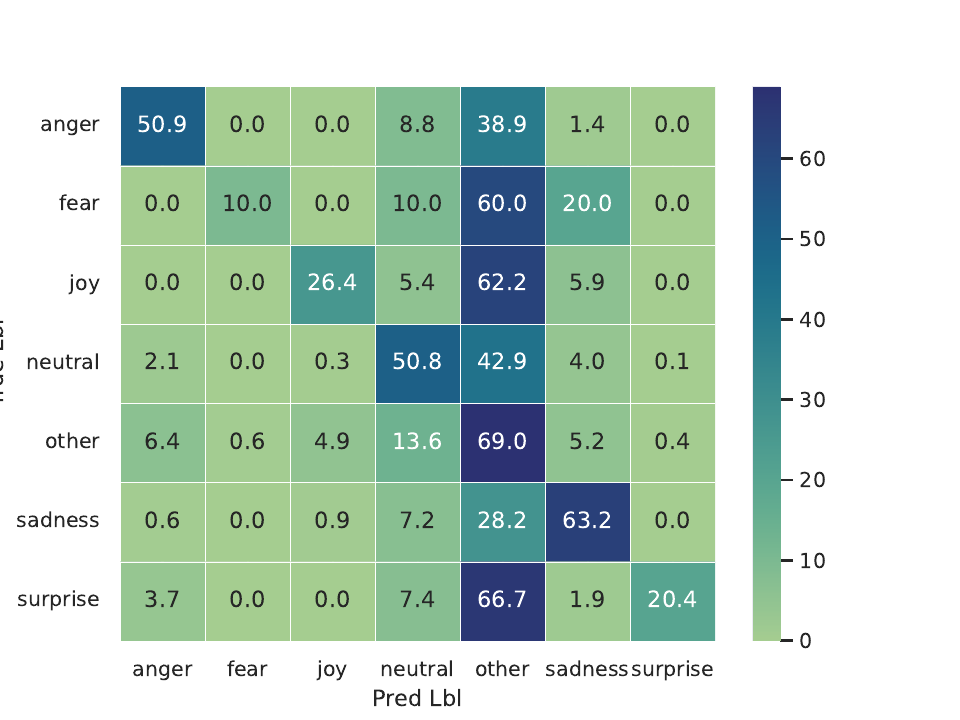}}
\subfloat[iemocap with INT (T5)]{\includegraphics[width=0.45\linewidth]{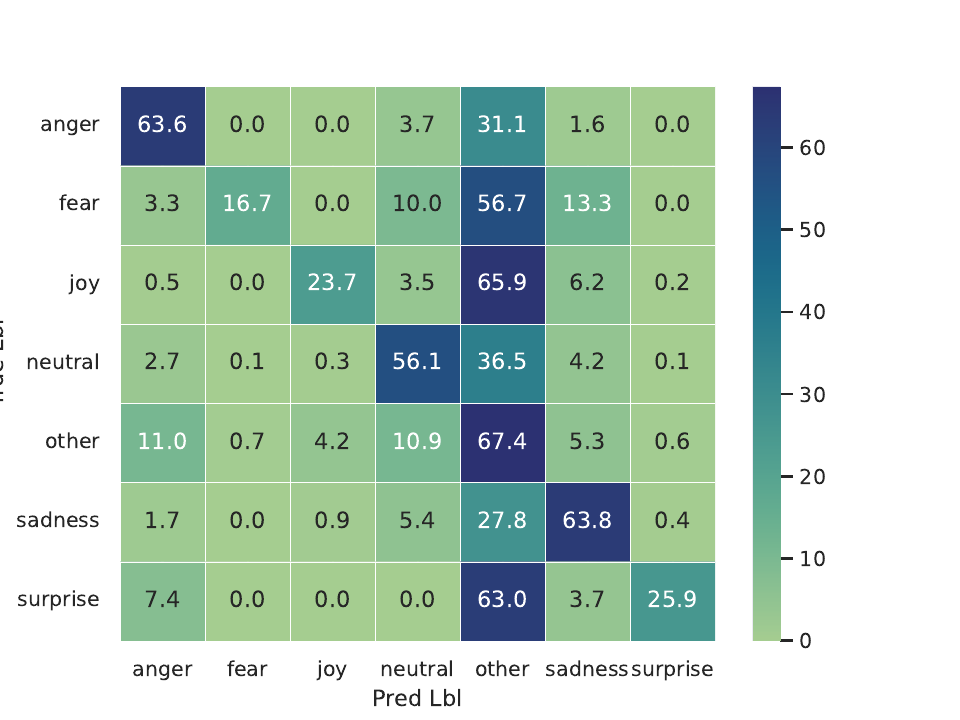}}
\vspace{-0.1cm}
\subfloat[CB with UTT (T5)]{\includegraphics[width=0.45\linewidth]{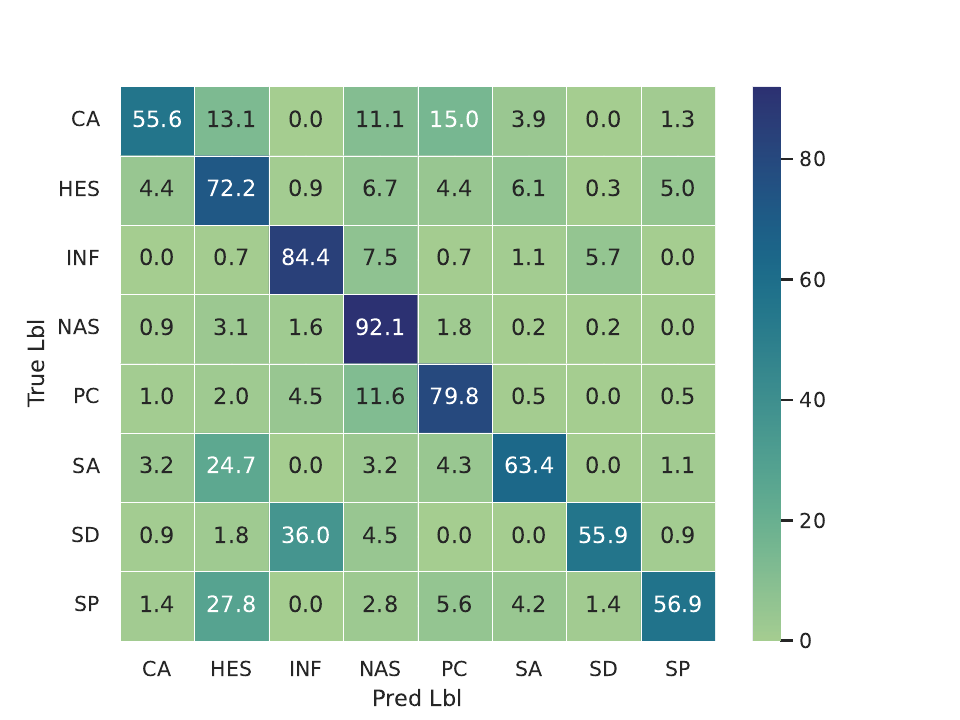}}
\subfloat[CB with ALL (T5)]{\includegraphics[width=0.45\linewidth]{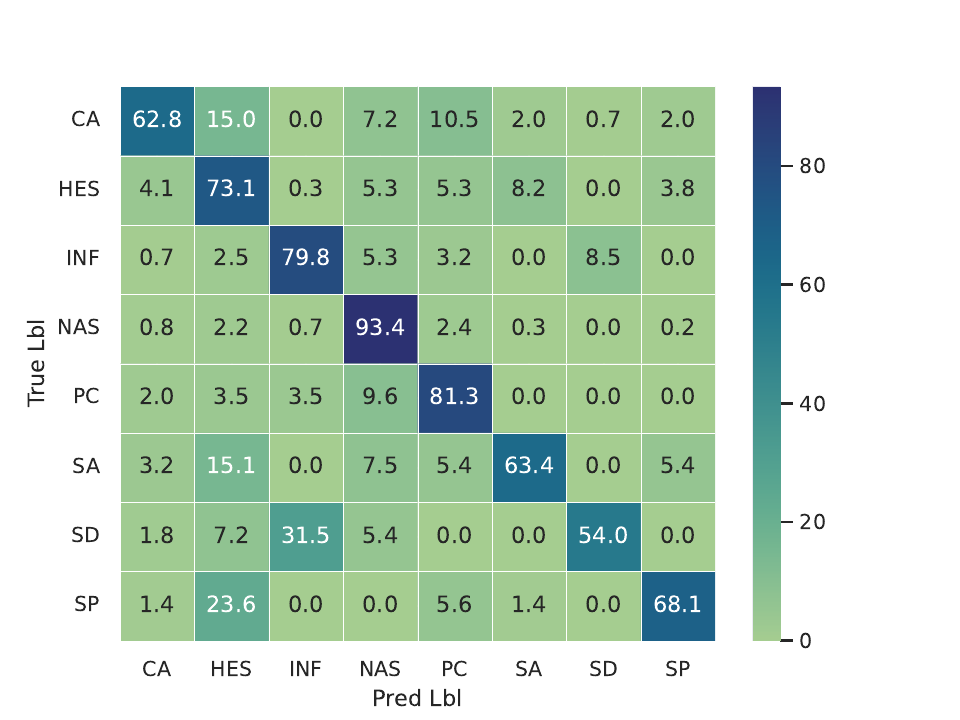}}
\vspace{-0.1cm}
\subfloat[P4G with UTT (T5)]{\includegraphics[width=0.45\linewidth]{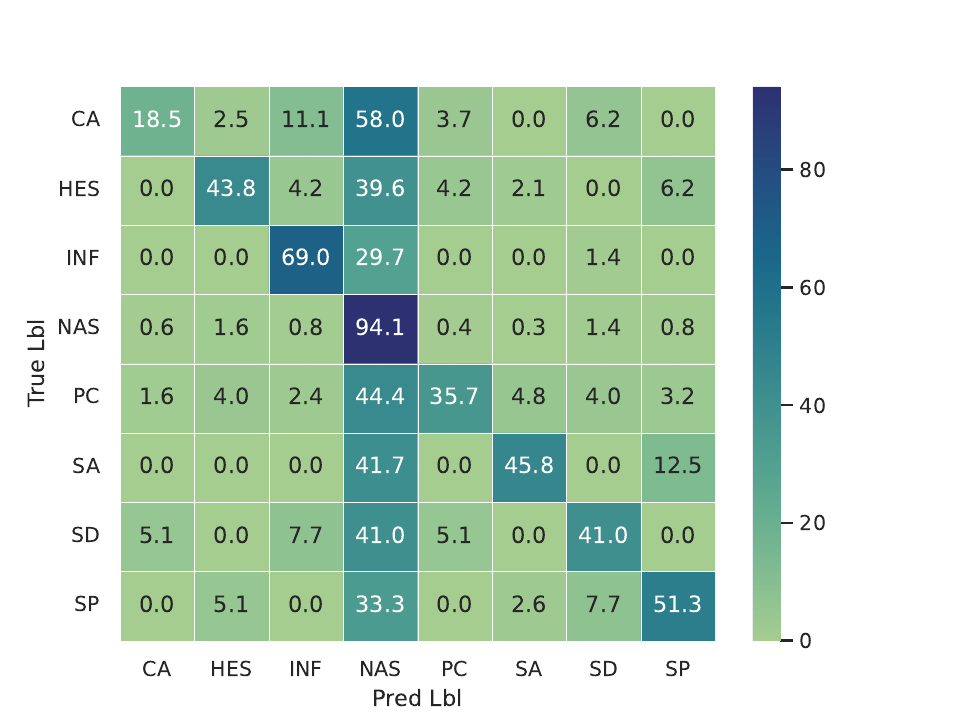}}
\subfloat[P4G with ALL (T5)]{\includegraphics[width=0.45\linewidth]{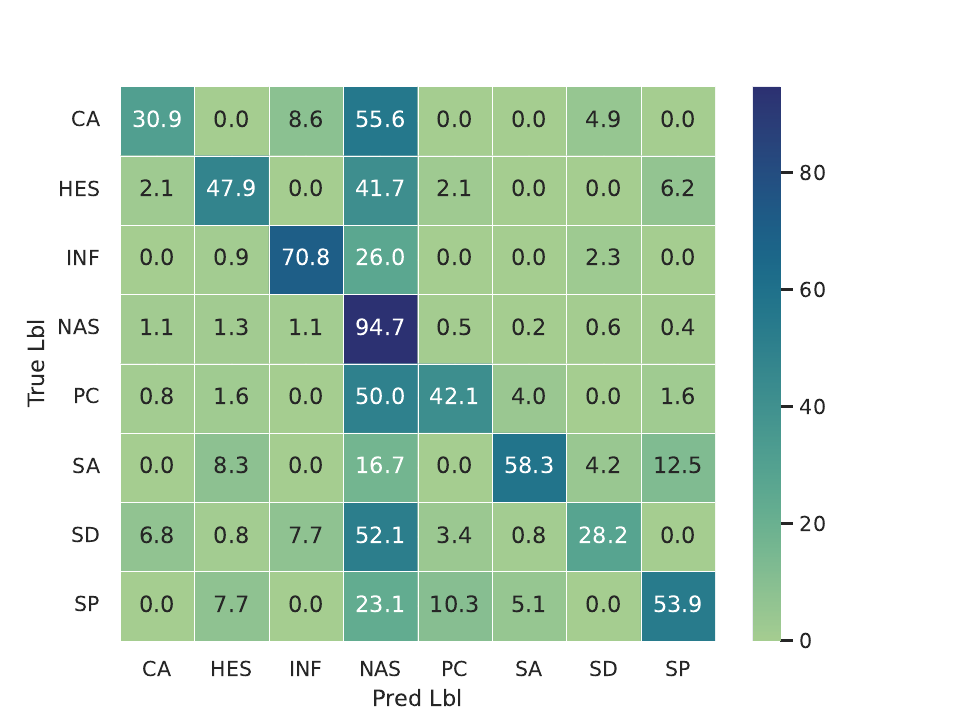}}
\caption{We present here the confusion matrices of the best performing pair of models and rationales in the in-domain setting for the 4 datasets and the corresponding model in absence of any rationale (UTT) in the in-domain setting (ID) }
\label{fig: ID-cm}
\end{figure*}

\begin{figure*}[]
\centering
\subfloat[friends with UTT (T5)]{\includegraphics[width=0.43\linewidth]{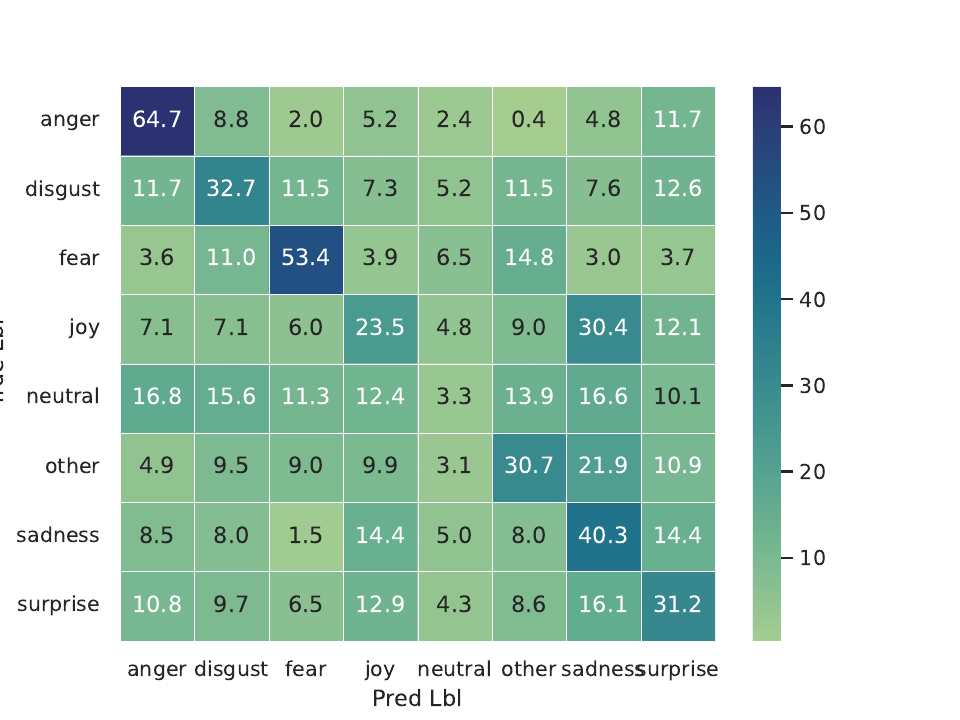}}
\subfloat[friends with ALL (T5)]{\includegraphics[width=0.43\linewidth]{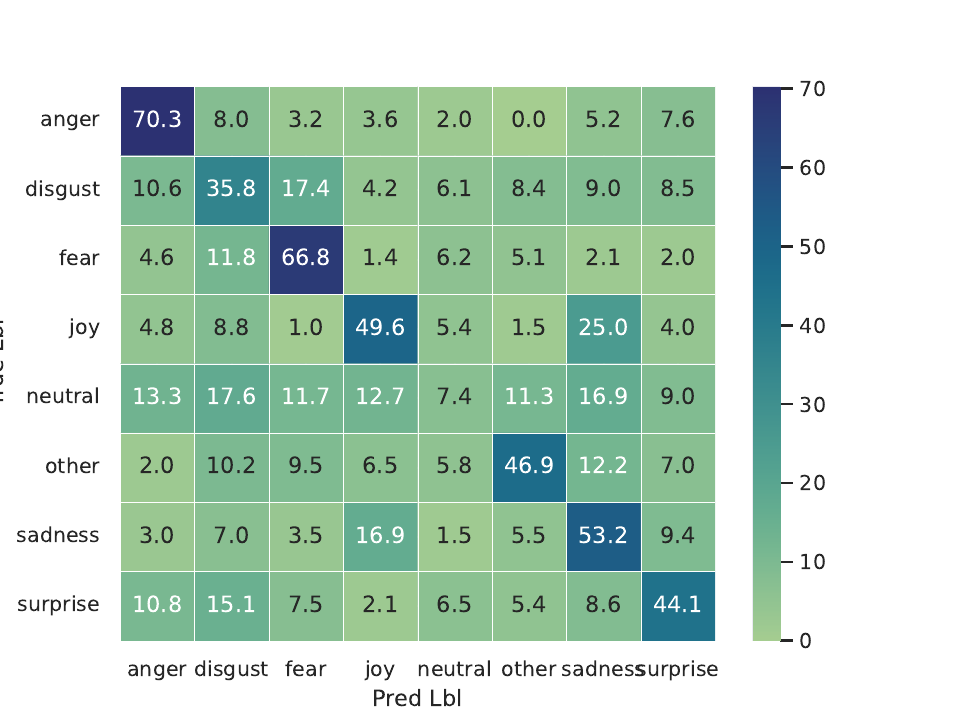}}
\vspace{-0.1cm}

\subfloat[iemocap with UTT (BERT)]{\includegraphics[width=0.43\linewidth]{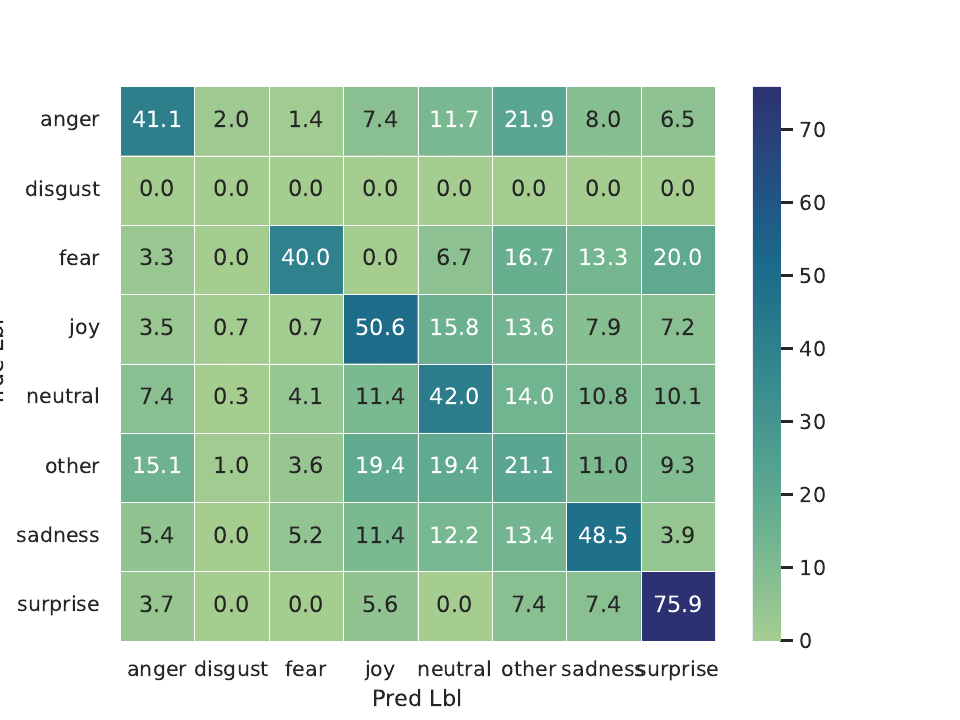}}
\subfloat[iemocap with INT (BERT)]{\includegraphics[width=0.43\linewidth]{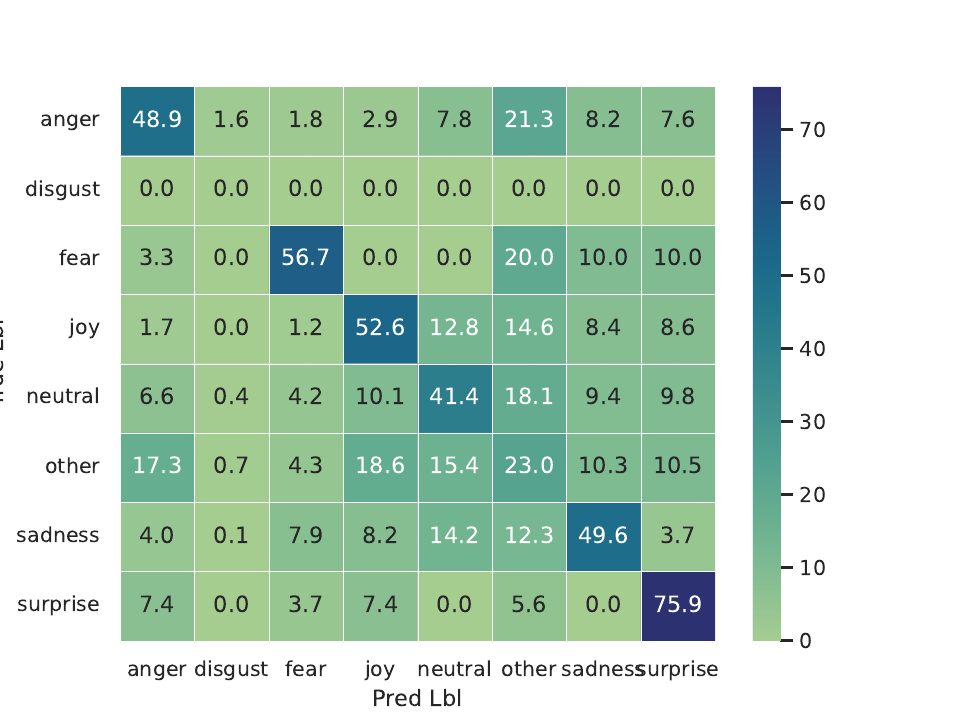}}
\vspace{-0.1cm}

\subfloat[CB with UTT (BERT)]{\includegraphics[width=0.43\linewidth]{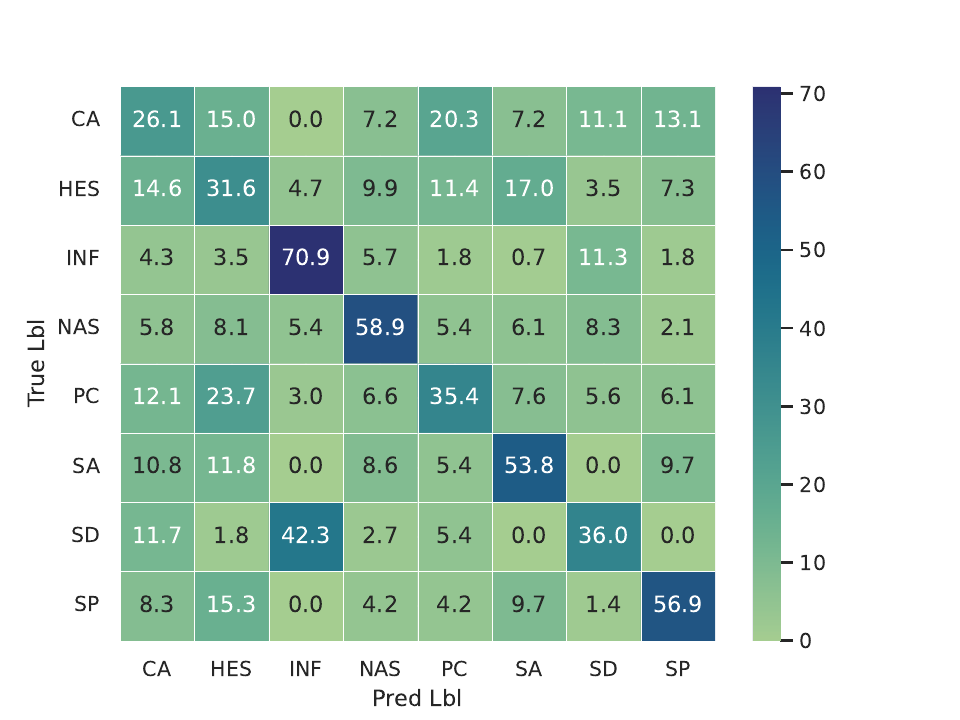}}
\subfloat[CB with ALL (BERT)]{\includegraphics[width=0.43\linewidth]{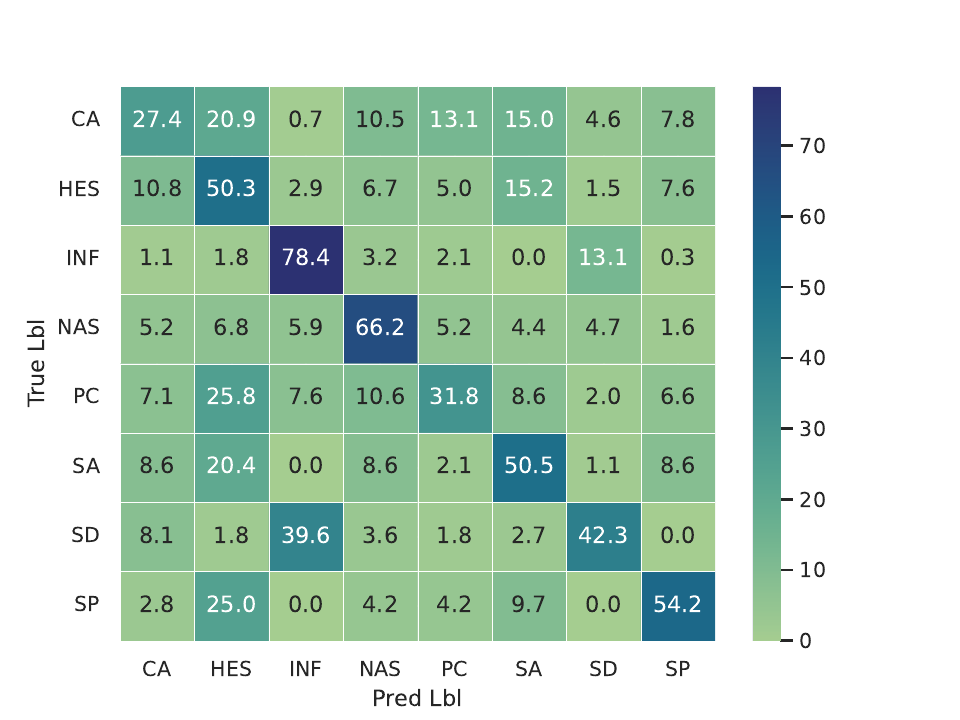}}
\vspace{-0.1cm}

\subfloat[P4G with UTT (BERT)]{\includegraphics[width=0.43\linewidth]{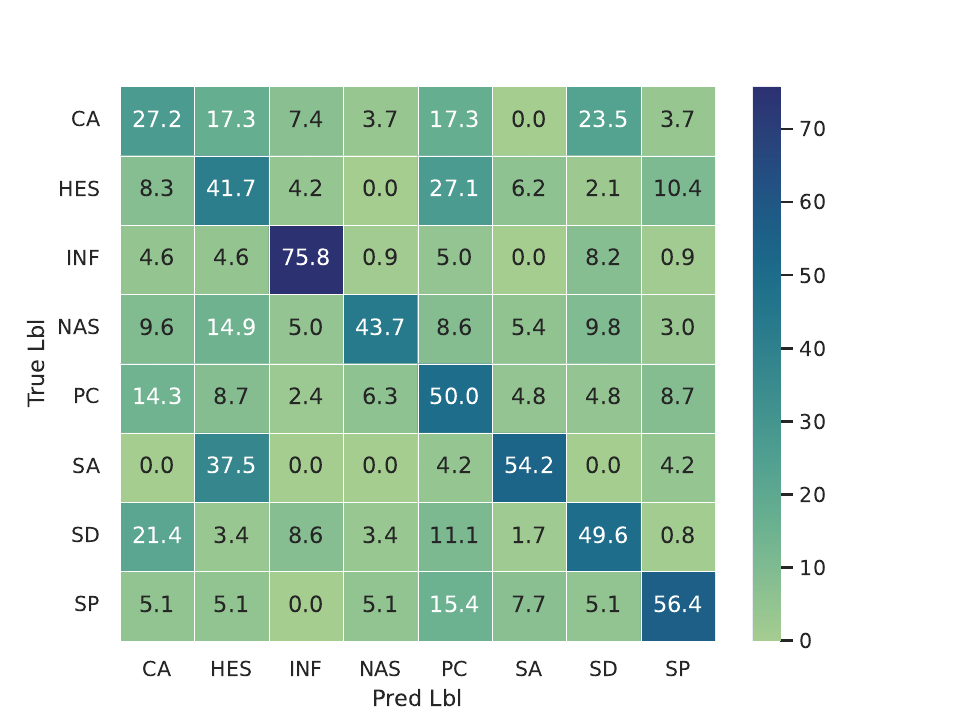}}
\subfloat[P4G with INT (BERT)]{\includegraphics[width=0.43\linewidth]{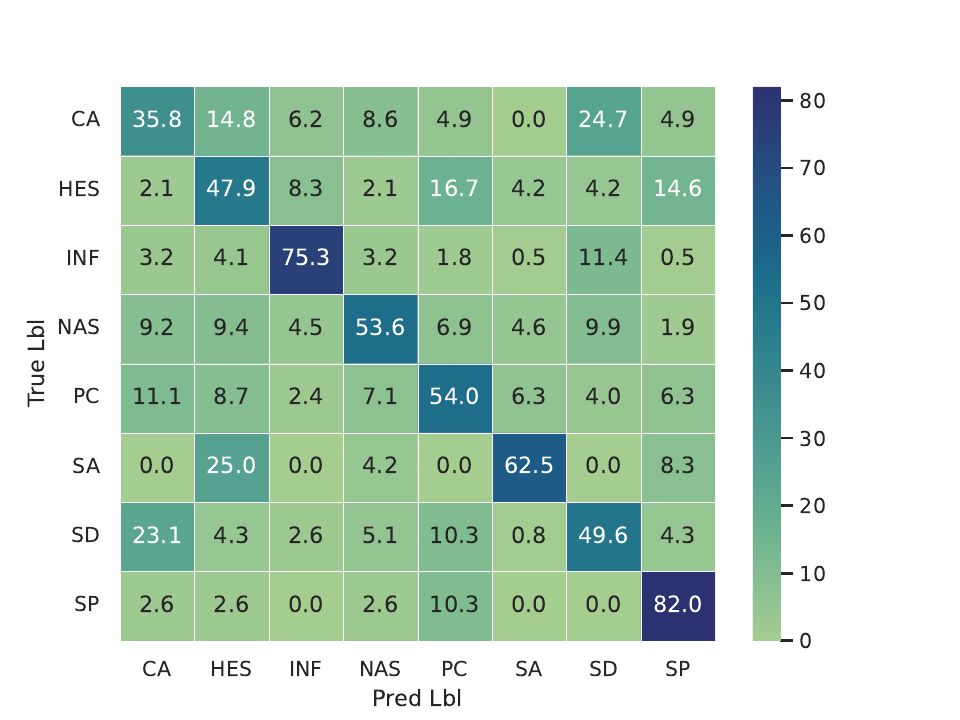}}

\caption{We present here the confusion matrices of the best performing pair of models and rationales in the transfer setting at k=20-shot case for the 4 datasets and the corresponding model in absence of any rationale (UTT). }
\label{fig: TF-cm}
\end{figure*}

\begin{figure*}[]
\centering
\subfloat[friends with ALL in ID (BERT)]{\includegraphics[width=0.45\linewidth]{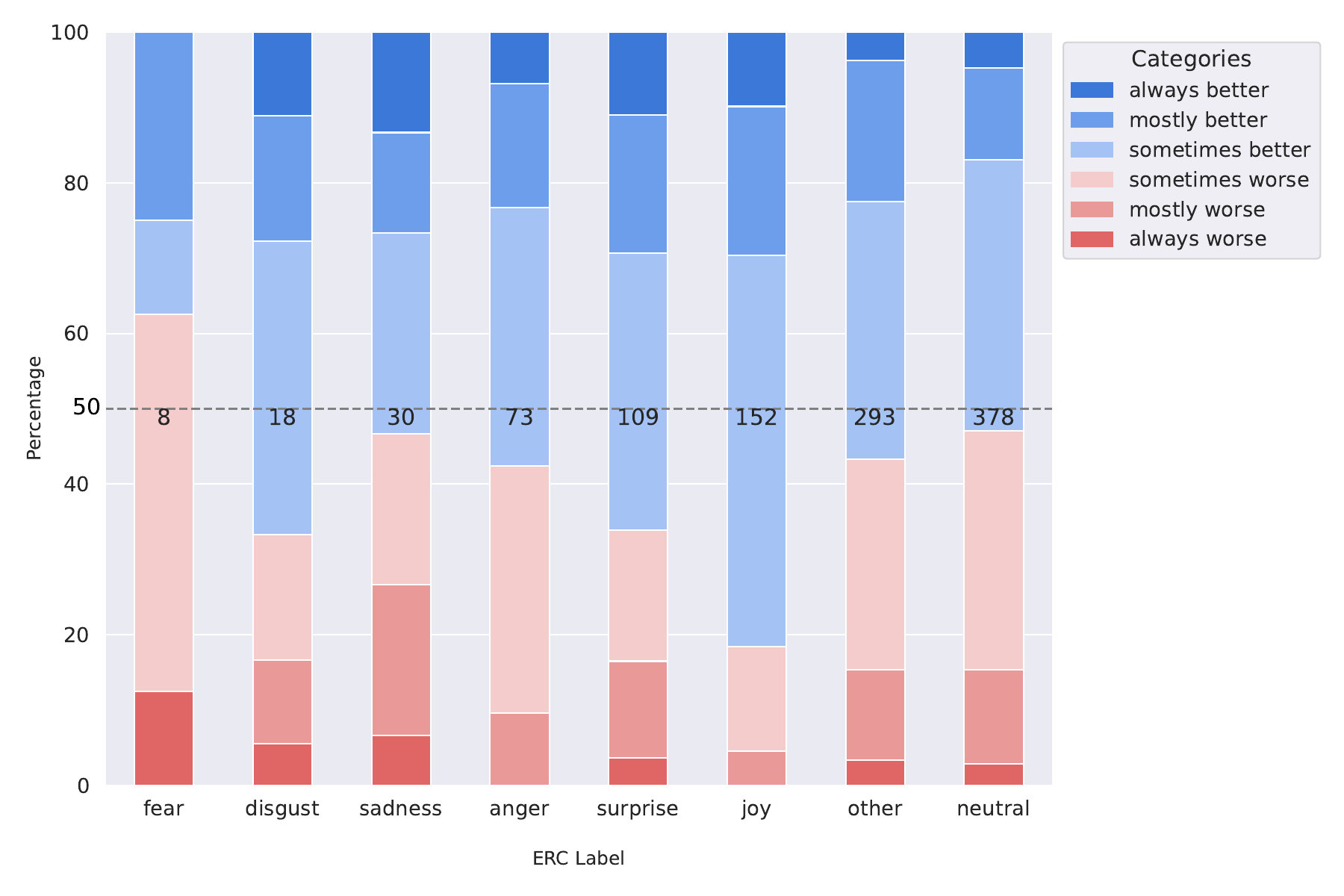}}
\subfloat[iemocap with INT in ID (T5)]{\includegraphics[width=0.45\linewidth]{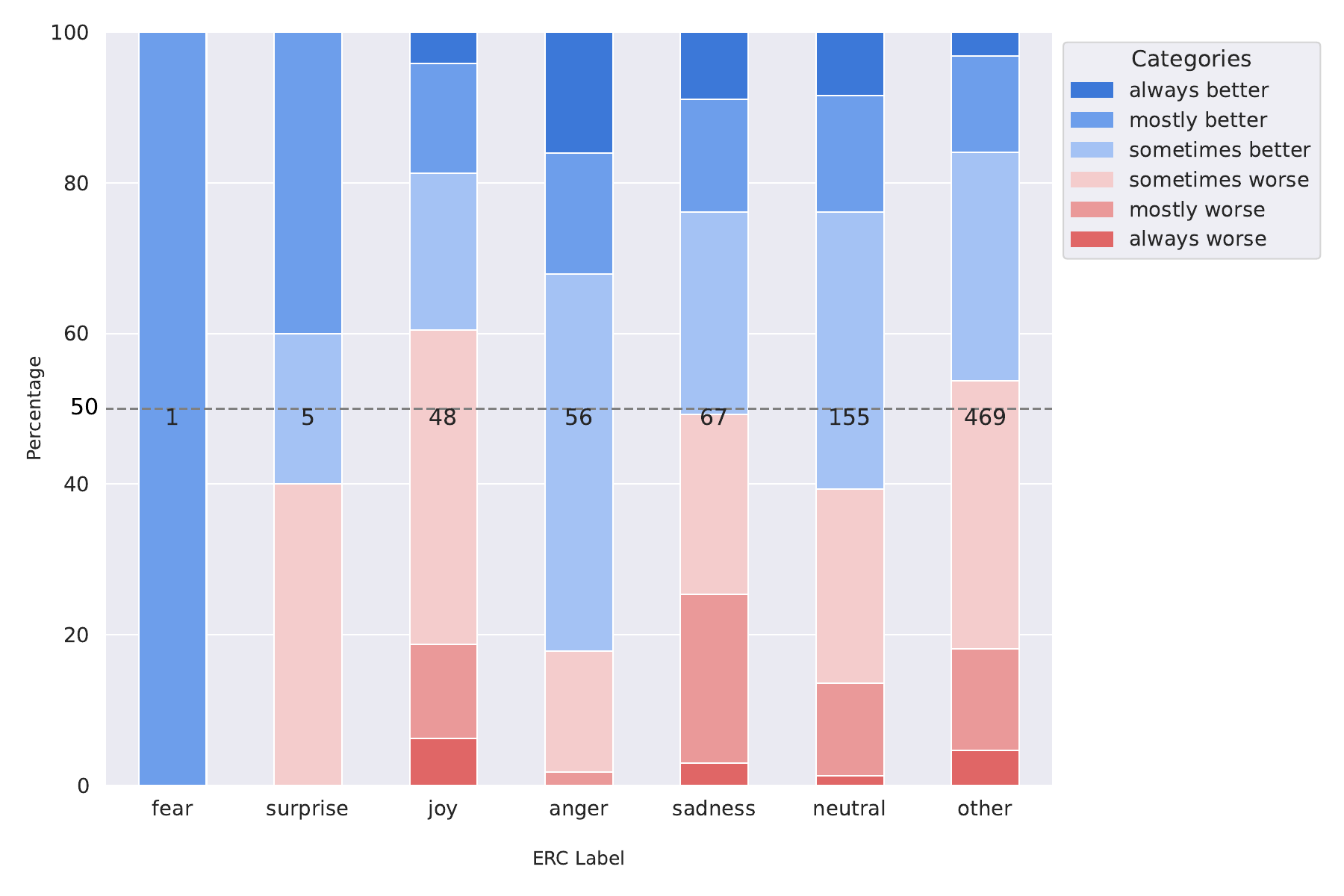}}

\subfloat[CB with ALL in ID (T5)]{\includegraphics[width=0.45\linewidth]{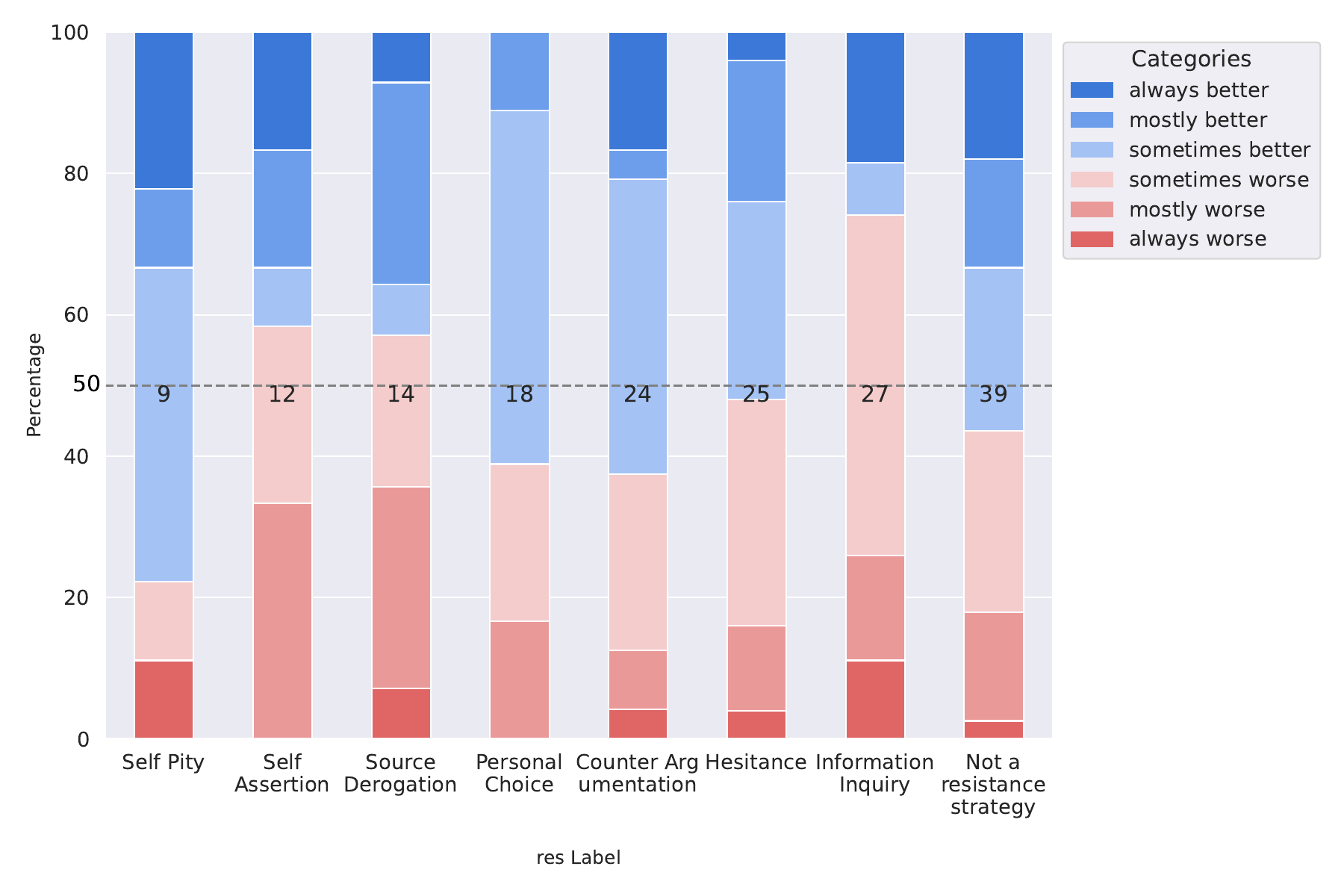}}
\subfloat[P4G with ALL in ID (T5)]{\includegraphics[width=0.45\linewidth]{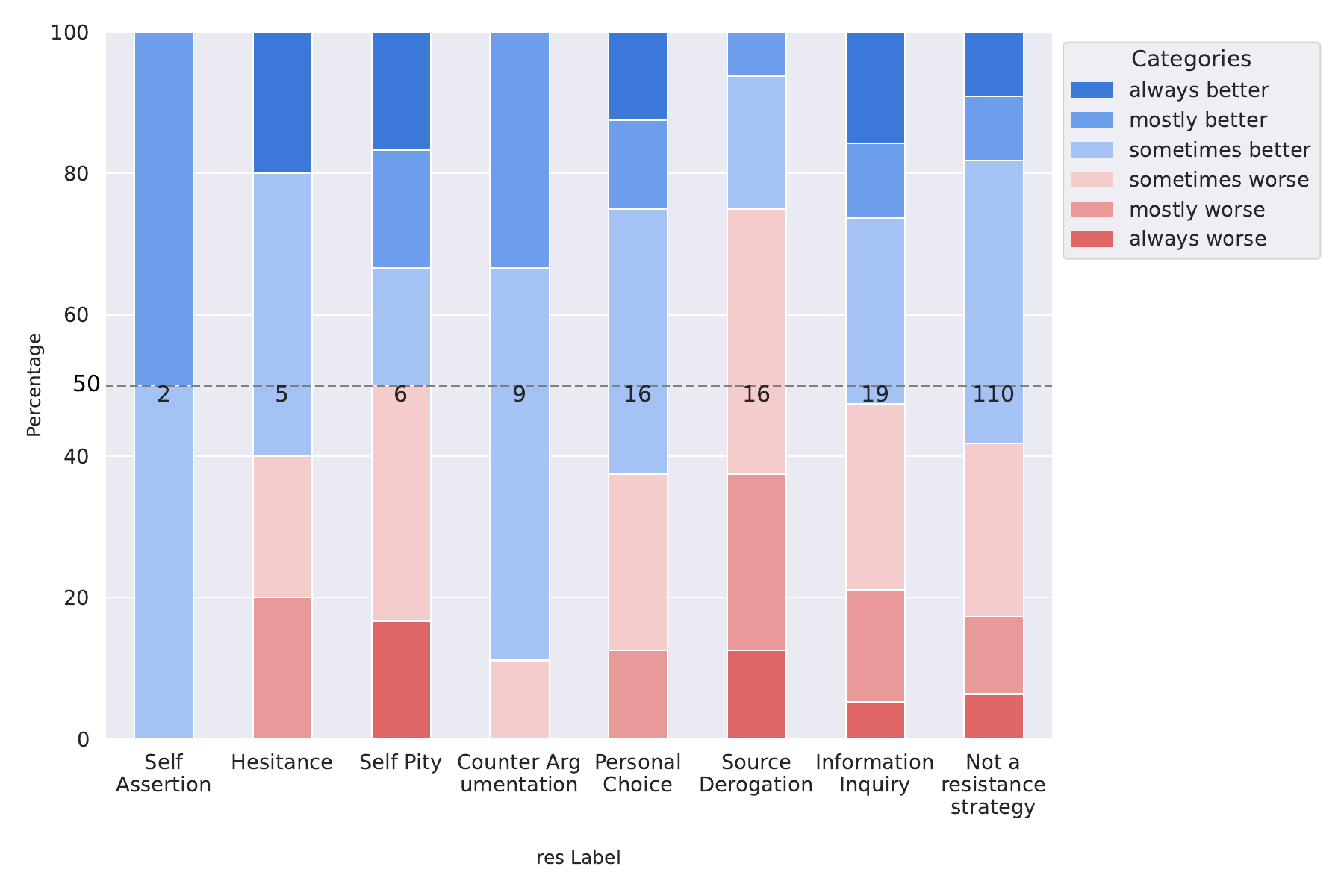}}

\subfloat[friends with ALL in 20-TF (T5)]{\includegraphics[width=0.45\linewidth]{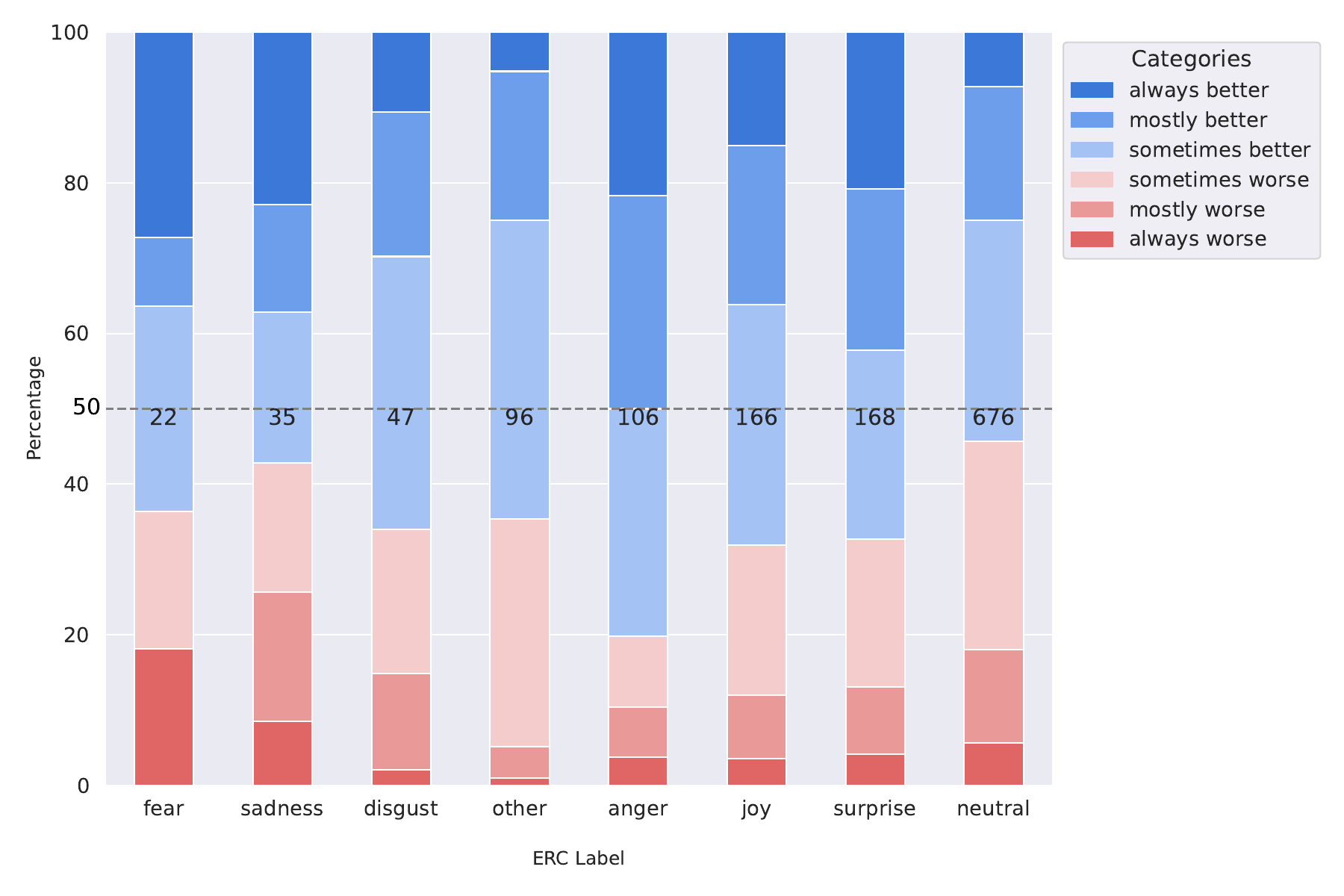}}
\subfloat[iemocap with ALL in 20-TF (BERT)]{\includegraphics[width=0.45\linewidth]{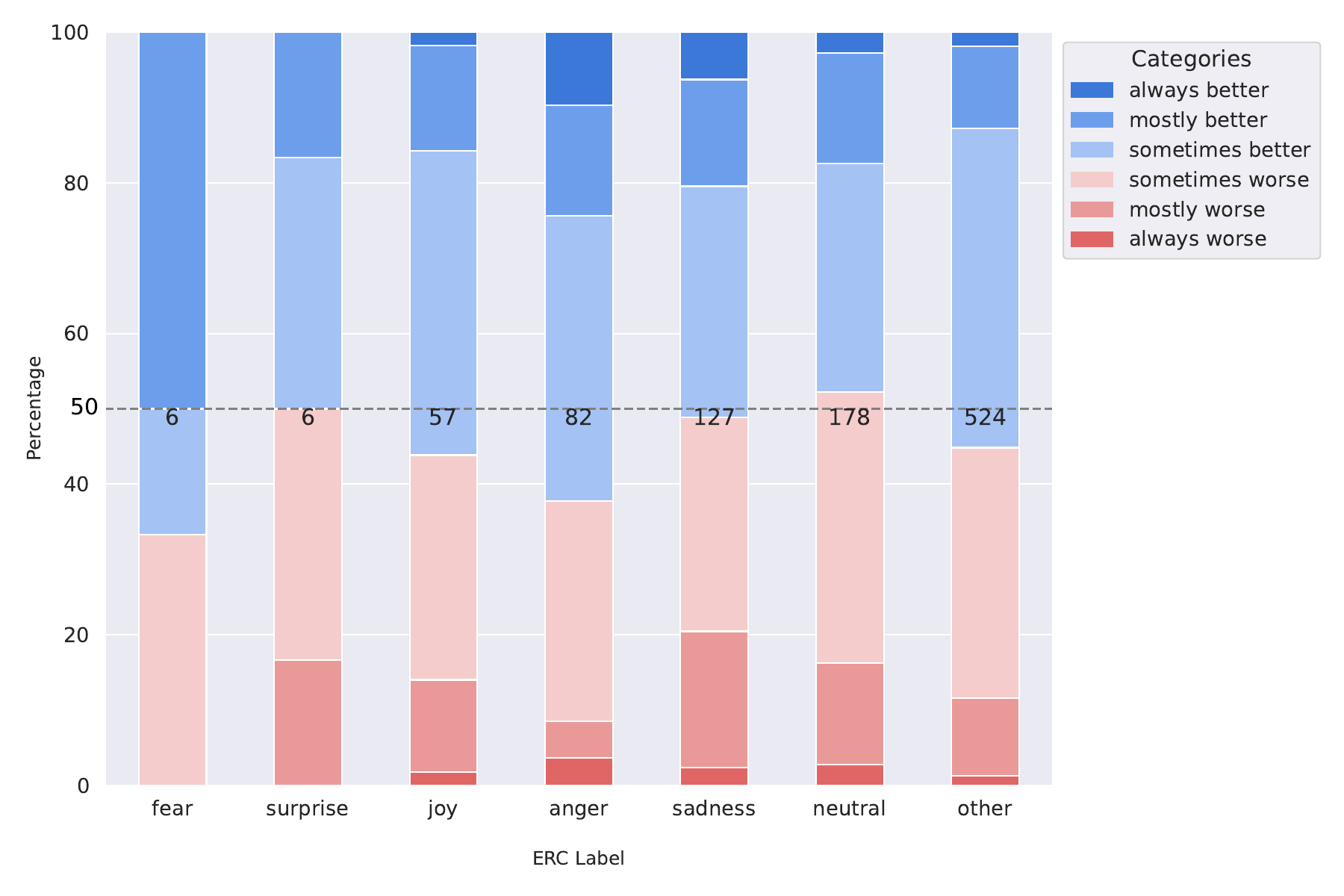}}

\subfloat[CB with ALL in 20-TF (BERT)]{\includegraphics[width=0.45\linewidth]{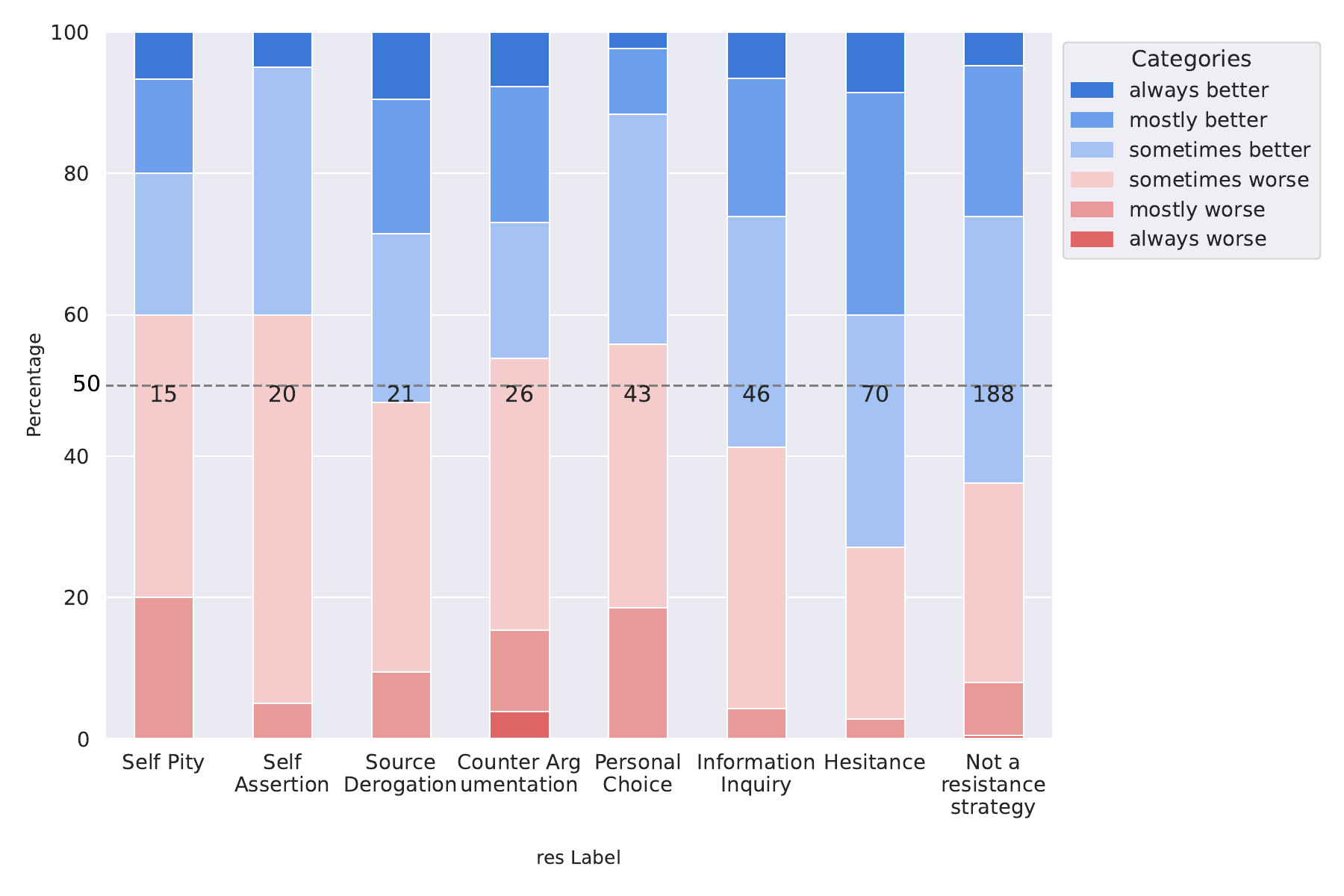}}
\subfloat[P4G with INT in 20-TF (BERT)]{\includegraphics[width=0.45\linewidth]{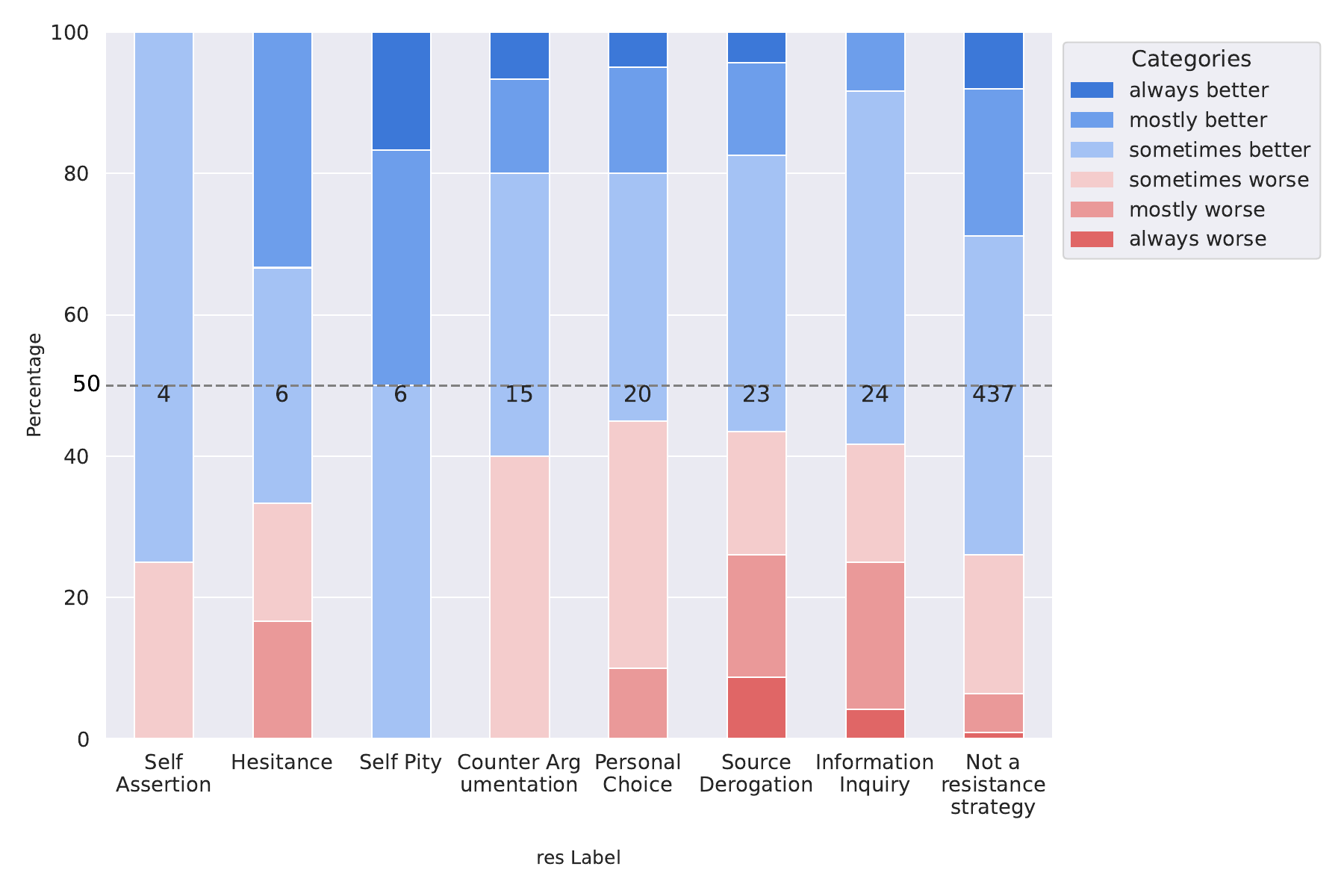}}

\caption{We present here the stacked bar plots that showcases the relative percentage of times a given label was predicted correctly by the best-performing model when augmented with a particular rationale as opposed to the baseline for different datasets. The labels are arranged in increasing order of frequency, with the number inside each bar indicating the frequency of the label. }
\label{fig: labelwise-difference}
\end{figure*}

We present the comprehensive results that could not make it into the main paper for lack of space, namely ID and TF results over all datasets with line-plots showing a direct visualization of the same in Figure \ref{fig:all-info-lineplot} and zero-shot resulys in Figure \ref{fig:zero-shot-perf}. We also highlight model mispredictions in terms of confusion matrices (Figures \ref{fig: ID-cm} and \ref{fig: TF-cm}) and highlight labels where models perform consistently better / worse  in Figure \ref{fig: labelwise-difference} and Tables \ref{table:utterance_analysis_better} and \ref{table:utterance_analysis_worse}.

\begin{table*}
\centering
\caption{Performance of different models on the \textbf{CB (Craigslist Bargain)} dataset for both in-domain (ID) and transfer (TF) setting across different few-shot splits (5, 10, 20, 50, 100) and the entire dataset (denoted by ``All''). The different rationales explored in this work are denoted by only utterance (-), utterance with speaker's intention (INT), utterance with the hearer's assumption (ASM), utterance with implicit information (IMP), and utterance with all the aforementioned rationales included i.e. INT, ASM, and IMP, and is denoted by ALL.}

\begin{tabular}{c c c c c c c c c}
\toprule
\textbf{Model} & \textbf{Mode} & \textbf{Rationale} & \textbf{5} & \textbf{10} & \textbf{20} & \textbf{50} & \textbf{100} & \textbf{All} \\
\midrule
\multirow{3}{*}{bert} & \multirow{3}{*}{ID} & - & 13.8±4.7 & 20.2±1.4 & 27.2±6.8 & 44.7±2.4 & 57.2±2.0 & 66.7±3.6 \\
 &  & INT & 13.4±5.5 & 22.9±0.7 & 34.6±3.4 & 50.3±2.5 & 59.3±1.8 & \textbf{68.4±1.7} \\
 &  & ASM & 13.0±5.9 & 22.3±5.0 & 30.4±1.7 & 47.2±1.4 & \textbf{60.4±3.0} & 66.6±0.7 \\
\multirow{7}{*}{} & \multirow{2}{*}{} & IMP & 13.6±5.0 & 23.8±3.1 & 31.6±6.7 & 50.9±2.5 & 60.1±1.9 & 66.9±0.3 \\
 &  & ALL & \textbf{16.0±6.4} & \textbf{24.8±4.4} & \textbf{38.8±2.2} & \textbf{51.6±1.2} & 58.5±1.9 & 67.0±0.7 \\
 \midrule
 & \multirow{5}{*}{TF} & - & 33.5±2.1 & 38.1±3.0 & 39.6±3.2 & 44.2±3.3 & 53.8±1.2 & 65.7±0.9 \\
 &  & INT & 35.3±0.4 & 41.0±4.0 & 42.2±1.9 & 49.4±2.9 & 56.3±1.2 & 64.8±3.5 \\
 &  & ASM & 35.1±0.8 & 39.7±2.4 & 41.7±1.4 & 48.3±1.8 & 54.3±2.1 & \textbf{66.8±0.6} \\
 &  & IMP & \textbf{37.5±1.4} & 42.4±1.6 & 42.8±0.5 & 50.2±4.5 & 55.0±1.9 & 66.1±2.9 \\
 &  & ALL & 37.1±1.9 & \textbf{44.5±2.9} & \textbf{44.8±0.7} & \textbf{52.4±2.0} & \textbf{57.9±0.6} & 66.3±1.6 \\
 \midrule
 
\multirow{3}{*}{gpt2} & \multirow{3}{*}{ID} & - & 7.6±6.3 & 12.1±5.7 & \textbf{20.6±4.4} & 30.7±6.3 & 43.0±1.0 & 60.0±0.9 \\
 &  & INT & 5.6±1.7 & \textbf{12.7±5.1} & 17.6±2.8 & \textbf{36.4±5.4} & 46.1±0.2 & 65.6±2.0 \\
 &  & ASM & 9.8±5.1 & 12.6±3.1 & 14.8±3.7 & 30.1±0.2 & 43.5±3.5 & 65.3±1.3 \\
\multirow{7}{*}{} & \multirow{2}{*}{} & IMP & 6.3±2.8 & 11.3±5.1 & 19.9±5.9 & 35.5±2.8 & \textbf{48.2±4.3} & 64.9±1.6 \\
 &  & ALL & \textbf{10.6±6.1} & 12.1±5.8 & 20.2±4.5 & 34.7±3.5 & 47.6±2.5 & \textbf{66.0±1.5} \\
 \midrule
 
 & \multirow{5}{*}{TF} & - & 26.9±2.4 & 31.8±1.3 & 32.7±2.3 & 38.1±0.6 & 43.0±2.7 & 60.4±0.4 \\
 &  & INT & \textbf{33.7±7.4} & \textbf{38.4±1.7} & \textbf{41.7±3.1} & \textbf{48.9±0.9} & 53.0±0.4 & 63.7±3.2 \\
 &  & ASM & 25.6±6.4 & 33.1±1.8 & 34.9±2.6 & 46.2±1.9 & 52.1±1.0 & 63.4±2.9 \\
 &  & IMP & 33.6±7.3 & 35.8±4.2 & 39.8±2.7 & 48.0±4.6 & \textbf{53.8±3.0} & 64.0±1.1 \\
 &  & ALL & 31.3±4.8 & 37.0±5.0 & 38.1±3.4 & 47.4±1.8 & 53.4±1.8 & \textbf{67.0±2.8} \\
 \midrule
 
\multirow{3}{*}{t5-base} & \multirow{3}{*}{ID} & - & 8.5±3.2 & 11.7±0.6 & \textbf{11.6±1.3} & 10.6±3.2 & 23.1±11.4 & 70.8±1.8 \\
 &  & INT & 7.3±2.5 & \textbf{11.8±1.9} & 11.5±1.7 & 10.7±3.4 & 30.7±1.6 & 70.6±2.8 \\
 &  & ASM & 9.2±2.6 & 7.9±0.9 & 11.2±2.3 & 7.0±0.3 & 23.4±3.1 & 69.0±1.8 \\
\multirow{7}{*}{} & \multirow{2}{*}{} & IMP & 8.0±4.3 & 7.8±1.6 & 11.3±1.1 & \textbf{10.8±3.0} & 29.8±2.8 & 69.1±2.6 \\
 &  & ALL & \textbf{9.4±2.5} & 8.2±2.2 & 10.1±1.5 & 10.4±4.0 & \textbf{37.7±3.9} & \textbf{72.2±0.5} \\
 \midrule
 
 & \multirow{5}{*}{TF} & - & 34.7±1.8 & 38.4±2.0 & 40.0±1.1 & 46.5±4.4 & \textbf{55.8±1.8} & 70.1±2.9 \\
 &  & INT & \textbf{34.9±4.3} & 38.1±2.2 & 41.3±3.5 & 53.1±0.8 & 55.3±3.3 & \textbf{72.1±0.7} \\
 &  & ASM & 33.9±3.0 & 38.9±0.5 & 42.5±2.6 & 50.2±3.0 & 53.0±3.1 & 70.3±3.4 \\
 &  & IMP & 28.5±2.2 & 37.8±3.1 & 39.6±5.3 & 45.7±0.8 & 50.7±1.6 & 70.5±1.3 \\
 &  & ALL & 33.2±4.5 & \textbf{39.4±2.0} & \textbf{43.7±2.2} & \textbf{50.3±3.9} & 54.6±3.7 & 69.6±1.7 \\
\bottomrule
\end{tabular}

\end{table*}


\begin{table*}
\centering

\caption{Performance of different models on the \textbf{P4G (Persuasion for Good)} dataset for both in-domain (ID) and transfer (TF) setting across different few-shot splits (5, 10, 20, 50, 100) and the entire dataset (denoted by ``All''). The different rationales explored in this work are denoted by only utterance (-), utterance with speaker's intention (INT), utterance with the hearer's assumption (ASM), utterance with implicit information (IMP), and utterance with all the aforementioned rationales included i.e. INT, ASM, and IMP, and is denoted by ALL.}

\begin{tabular}{c c c c c c c c c}
\toprule
\textbf{Model} & \textbf{Mode} & \textbf{Rationale} & \textbf{5} & \textbf{10} & \textbf{20} & \textbf{50} & \textbf{100} & \textbf{ALL} \\
\midrule
\multirow{3}{*}{bert} & \multirow{3}{*}{ID} & - & 9.6±0.2 & 13.5±3.8 & 19.8±0.6 & 29.2±0.7 & 32.9±1.1 & 50.6±2.5 \\
 &  & INT & \textbf{10.4±5.0} & \textbf{17.9±2.9} & 22.4±3.6 & 31.5±1.4 & 34.2±1.8 & 53.0±1.6 \\
 &  & ASM & 6.3±3.2 & 16.4±1.6 & 17.2±5.8 & 32.1±0.5 & 34.3±1.4 & 49.4±8.1 \\
\multirow{7}{*}{} & \multirow{2}{*}{} & IMP & 6.8±4.6 & 15.1±2.2 & 22.0±1.9 & 32.2±0.7 & 35.5±1.5 & 52.3±1.7 \\
 &  & ALL & 8.4±7.5 & 15.3±6.6 & \textbf{23.0±1.1} & \textbf{32.9±0.7} & \textbf{36.7±1.0} & \textbf{53.2±1.4} \\
 \midrule
 & \multirow{5}{*}{TF} & - & 22.7±0.3 & 23.9±0.9 & 26.7±1.5 & 29.5±2.6 & 32.2±0.4 & 48.4±1.4 \\
 &  & INT & 26.4±1.1 & \textbf{29.0±2.7} & \textbf{32.2±1.0} & 33.7±0.4 & 35.6±2.0 & 49.0±0.6 \\
 &  & ASM & 24.4±2.8 & 26.2±2.0 & 26.9±1.0 & 30.0±0.6 & 33.0±1.6 & 47.0±3.5 \\
 &  & IMP & 22.2±3.3 & 25.1±2.3 & 28.0±1.2 & 32.4±0.6 & 34.2±1.5 & 48.2±0.7 \\
 &  & ALL & \textbf{27.0±0.9} & 29.0±2.2 & 31.1±0.7 & \textbf{34.9±2.0} & \textbf{37.5±2.3} & \textbf{50.2±3.5} \\
 \midrule
\multirow{3}{*}{gpt2} & \multirow{3}{*}{ID} & - & 4.2±2.5 & 6.3±4.1 & \textbf{10.0±3.2} & 16.5±1.1 & 19.2±1.9 & 35.7±4.4 \\
 &  & INT & 3.7±2.4 & \textbf{6.9±2.7} & 7.9±3.0 & 20.6±1.4 & 26.3±3.3 & 45.7±1.6 \\
 &  & ASM & 2.7±1.1 & 3.7±1.2 & 7.3±1.9 & 16.2±5.3 & 28.4±5.7 & 47.7±2.4 \\
\multirow{7}{*}{} & \multirow{2}{*}{} & IMP & 3.9±2.5 & 6.4±3.5 & 8.2±4.5 & 20.2±5.5 & 27.0±2.6 & 50.1±2.6 \\
 &  & ALL & 2.1±0.8 & 6.4±1.9 & 9.0±3.9 & \textbf{21.8±1.7} & \textbf{29.0±4.5} & \textbf{50.1±1.4} \\
 \midrule
 & \multirow{5}{*}{TF} & - & 13.5±1.8 & 16.3±0.5 & 16.6±2.9 & 19.0±0.6 & 21.3±1.3 & 32.4±3.6 \\
 &  & INT & 18.3±2.0 & 20.7±0.4 & 24.6±0.6 & 28.5±2.1 & 30.2±0.3 & 46.4±1.8 \\
 &  & ASM & 20.4±1.2 & 21.0±1.1 & 23.6±1.5 & 26.6±1.9 & 29.6±0.9 & 45.0±2.0 \\
 &  & IMP & 18.8±3.6 & 22.4±1.8 & 23.9±1.6 & 29.1±1.7 & 29.7±2.3 & \textbf{48.4±2.1} \\
 &  & ALL & \textbf{20.6±2.7} & \textbf{23.6±0.3} & \textbf{25.5±2.2} & \textbf{30.6±0.2} & \textbf{31.5±2.0} & 47.5±2.0 \\
 \midrule
\multirow{3}{*}{t5-base} & \multirow{3}{*}{ID} & - & 10.3±0.9 & 12.6±2.6 & 6.5±2.3 & 8.3±2.6 & 10.5±1.9 & 48.8±0.9 \\
 &  & INT & 10.4±1.0 & 9.2±5.6 & 6.6±0.2 & 10.0±0.8 & 12.7±0.7 & 51.2±1.4 \\
 &  & ASM & 11.7±2.1 & 10.2±4.3 & 8.7±3.9 & 6.8±0.8 & 12.0±3.2 & 51.1±0.8 \\
\multirow{7}{*}{} & \multirow{2}{*}{} & IMP & 11.1±1.8 & 7.7±3.5 & 7.0±2.7 & 8.0±4.2 & 11.7±8.9 & 51.7±3.0 \\
 &  & ALL & 13.4±1.1 & 10.7±4.6 & 7.4±3.9 & 7.7±1.4 & \textbf{22.4±7.5} & \textbf{53.4±2.7} \\
 \midrule
 & \multirow{5}{*}{TF} & - & 19.2±1.6 & 22.0±2.0 & 23.9±1.6 & 28.4±0.9 & 32.6±0.9 & 51.2±2.3 \\
 &  & INT & 19.9±3.5 & 24.1±2.1 & 25.9±2.6 & \textbf{33.5±2.6} & 31.4±4.4 & 51.3±1.6 \\
 &  & ASM & 19.6±2.0 & \textbf{24.7±3.9} & 26.0±1.3 & 29.1±1.3 & 32.6±1.6 & 49.3±0.8 \\
 &  & IMP & \textbf{21.5±1.5} & 24.4±0.5 & \textbf{29.1±1.8} & 30.9±1.0 & 33.5±3.6 & 51.4±2.9 \\
 &  & ALL & 19.2±2.1 & 21.3±1.7 & 28.0±2.8 & 30.8±3.0 & \textbf{37.5±0.8} & \textbf{53.2±1.2} \\
\bottomrule
\end{tabular}

\end{table*}

\begin{table*}
\centering

\caption{Performance of different models on the \textbf{Friends} dataset for the task of ERC for both in-domain (ID) and transfer (TF) setting across different few-shot splits (5, 10, 20, 50, 100) and the entire dataset (denoted by ``All''). The different rationales explored in this work are denoted by only utterance (-), utterance with speaker's intention (INT), utterance with the hearer's assumption (ASM), utterance with implicit information (IMP), and utterance with all the aforementioned rationales included i.e. INT, ASM, and IMP, and is denoted by ALL.}

\begin{tabular}{c c c c c c c c c}
\toprule
\textbf{Model} & \textbf{Mode} & \textbf{Rationale} & \textbf{5} & \textbf{10} & \textbf{20} & \textbf{50} & \textbf{100} & \textbf{All} \\
\midrule
\multirow{3}{*}{bert} & \multirow{3}{*}{ID} & - & 13.4±2.1 & 15.0±2.1 & 17.5±3.7 & 31.2±0.8 & 33.9±0.7 & 40.9±0.9 \\
 &  & INT & 13.2±1.2 & \textbf{19.2±2.6} & \textbf{26.9±5.8} & \textbf{34.5±3.0} & \textbf{39.9±1.8} & 45.3±0.8 \\
 &  & ASM & 11.5±5.0 & 16.4±3.4 & 18.6±3.8 & 30.2±0.8 & 35.4±1.3 & 44.6±0.1 \\
\multirow{7}{*}{} & \multirow{2}{*}{} & IMP & 12.5±4.9 & 13.2±3.2 & 22.5±3.9 & 32.1±1.6 & 36.0±1.0 & 44.7±1.7 \\
 &  & ALL & 5.8±3.7 & 16.9±3.4 & 23.8±5.3 & 33.6±2.0 & 37.7±1.0 & \textbf{46.2±1.3} \\
 \midrule
 
 & \multirow{5}{*}{TF} & - & 22.3±1.1 & 24.7±0.8 & 26.3±2.1 & 29.2±2.0 & 31.6±1.6 & 41.0±1.3 \\
 &  & INT & 24.2±2.0 & \textbf{25.0±2.4} & \textbf{28.3±1.5} & \textbf{30.6±1.0} & 32.6±1.1 & 44.9±0.4 \\
 &  & ASM & 23.2±3.0 & 23.9±2.4 & 24.9±2.7 & 27.3±1.4 & 30.8±1.0 & 40.9±0.8 \\
 &  & IMP & 21.4±1.2 & 24.2±1.5 & 25.1±1.6 & 28.1±0.9 & 31.5±1.5 & 45.0±0.6 \\
 &  & ALL & \textbf{25.3±2.2} & 24.3±1.9 & 27.6±1.2 & 30.2±1.3 & \textbf{33.1±1.0} & \textbf{46.1±1.8} \\
 \midrule
 
\multirow{3}{*}{gpt2} & \multirow{3}{*}{ID} & - & 7.7±0.9 & 9.9±0.9 & 11.2±0.2 & 12.2±1.0 & 17.1±1.1 & 26.5±0.8 \\
 &  & INT & 7.3±1.8 & 9.5±0.3 & 10.0±1.5 & \textbf{23.6±2.1} & \textbf{28.4±3.2} & 44.5±1.0 \\
 &  & ASM & 5.7±0.8 & 7.6±0.5 & 10.0±1.4 & 14.0±1.3 & 20.6±3.4 & 43.4±1.2 \\
\multirow{7}{*}{} & \multirow{2}{*}{} & IMP & 7.9±2.4 & 9.0±1.1 & 10.1±0.9 & 15.2±1.7 & 24.0±1.4 & 43.3±1.9 \\
 &  & ALL & 7.9±1.1 & 8.8±0.4 & 10.6±3.6 & 18.5±1.3 & 27.1±1.6 & \textbf{45.5±0.8} \\
 \midrule
 
 & \multirow{5}{*}{TF} & - & 14.2±1.2 & 14.6±0.2 & 14.9±0.9 & 15.9±1.0 & 17.9±1.4 & 26.5±1.3 \\
 &  & INT & \textbf{21.5±2.6} & 22.0±1.1 & \textbf{27.2±1.5} & \textbf{27.9±0.8} & 30.8±1.5 & 43.7±1.7 \\
 &  & ASM & 14.5±3.1 & 16.4±3.8 & 18.7±0.9 & 20.6±1.6 & 26.3±1.8 & 40.7±0.7 \\
 &  & IMP & 16.9±2.3 & 16.9±3.3 & 20.6±1.6 & 23.2±1.5 & 27.7±2.5 & 42.6±1.1 \\
 &  & ALL & 19.9±3.1 & \textbf{22.5±1.5} & 26.5±0.9 & 27.5±1.8 & \textbf{31.0±2.9} & \textbf{45.4±1.1} \\
 \midrule
 
\multirow{3}{*}{t5-base} & \multirow{3}{*}{ID} & - & 4.8±4.3 & 11.5±0.3 & 12.3±1.4 & 16.2±3.8 & 25.5±0.4 & 39.8±3.4 \\
 &  & INT & 11.6±5.4 & \textbf{20.1±1.5} & 26.5±2.7 & 24.5±2.5 & 28.3±2.3 & \textbf{44.8±2.6} \\
 &  & ASM & 11.3±1.5 & 13.4±1.3 & 18.5±2.6 & 20.0±2.3 & 23.2±2.8 & 39.8±0.6 \\
\multirow{7}{*}{} & \multirow{2}{*}{} & IMP & 11.1±0.3 & 15.4±4.0 & 19.8±2.8 & 22.7±3.1 & 25.4±5.1 & 44.1±3.3 \\
 &  & ALL & \textbf{16.9±2.5} & 20.0±1.1 & \textbf{28.0±2.1} & \textbf{26.5±1.2} & \textbf{31.3±1.8} & 43.8±3.1 \\
 \midrule
 
 & \multirow{5}{*}{TF} & - & 19.0±0.5 & 20.4±1.3 & 21.4±1.7 & 26.1±2.5 & \textbf{31.2±1.3} & 40.3±2.9 \\
 &  & INT & 24.5±2.4 & \textbf{26.1±2.7} & 27.8±2.6 & 29.9±1.2 & 30.9±1.3 & 42.6±2.9 \\
 &  & ASM & 19.7±2.0 & 22.6±2.4 & 23.0±1.0 & 26.2±0.9 & 29.2±1.3 & 44.6±2.3 \\
 &  & IMP & 20.6±1.4 & 22.8±1.0 & 24.7±1.2 & 28.2±1.3 & 30.2±1.7 & 47.2±0.4 \\
 &  & ALL & \textbf{24.8±2.3} & 25.0±1.1 & \textbf{28.0±1.5} & \textbf{30.0±0.9} & 30.7±0.7 & \textbf{47.4±0.7} \\
 \bottomrule
\end{tabular}

\end{table*}

\begin{table*}
\centering
\caption{Performance of different models on the \textbf{IEMOCAP} dataset for the task of ERC for both in-domain (ID) and transfer (TF) setting across different few-shot splits (5, 10, 20, 50, 100) and the entire dataset (denoted by ``All''). The different rationales explored in this work are denoted by only utterance (-), utterance with speaker's intention (INT), utterance with the hearer's assumption (ASM), utterance with implicit information (IMP), and utterance with all the aforementioned rationales included i.e. INT, ASM, and IMP, and is denoted by ALL.}

\begin{tabular}{c c c c c c c c c}
\toprule
\textbf{Model} & \textbf{Mode} & \textbf{Rationale} & \textbf{5} & \textbf{10} & \textbf{20} & \textbf{50} & \textbf{100} & \textbf{all} \\
\midrule
\multirow{3}{*}{bert} & \multirow{3}{*}{ID} & - & 13.7±7.2 & 16.1±3.1 & 24.3±2.7 & 33.0±1.4 & 36.1±1.1 & 40.7±1.5 \\
 &  & INT & 11.6±4.6 & 14.8±0.7 & 23.2±1.2 & 35.0±1.5 & \textbf{38.3±1.5} & \textbf{42.6±1.3} \\
 &  & ASM & 10.8±5.2 & \textbf{19.6±2.7} & 22.0±1.4 & 32.8±1.5 & 35.8±4.3 & 41.0±1.8 \\
\multirow{7}{*}{} & \multirow{2}{*}{} & IMP & 13.3±1.7 & 14.4±5.6 & \textbf{25.2±1.6} & 32.2±3.4 & 36.3±2.7 & 42.0±1.2 \\
 &  & ALL & 12.1±5.4 & 15.7±2.8 & 25.0±1.4 & \textbf{35.5±2.6} & 37.6±1.5 & 40.4±1.0 \\
 \midrule
 & \multirow{5}{*}{TF} & - & 23.6±1.9 & 23.8±3.4 & 24.0±2.4 & 27.1±1.0 & 29.5±0.4 & 37.1±0.8 \\
 &  & INT & 22.8±2.2 & 23.6±1.8 & 24.3±1.3 & \textbf{29.0±2.4} & 30.4±0.9 & \textbf{43.4±1.5} \\
 &  & ASM & 23.8±1.0 & 24.2±0.5 & 24.4±1.0 & 26.9±2.5 & \textbf{32.5±4.0} & 39.4±2.4 \\
 &  & IMP & 25.0±1.0 & 24.6±1.7 & \textbf{25.9±1.3} & 27.0±1.4 & 29.9±0.3 & 42.1±0.9 \\
 &  & ALL & \textbf{25.4±0.4} & \textbf{25.0±1.8} & 25.6±0.7 & 28.3±0.5 & 30.6±1.3 & 40.7±5.3 \\
 \midrule
\multirow{3}{*}{gpt2} & \multirow{3}{*}{ID} & - & 5.0±4.2 & 6.0±4.7 & \textbf{10.6±2.2} & \textbf{17.1±2.2} & 23.4±3.0 & 35.3±2.4 \\
 &  & INT & 5.0±3.3 & 8.8±2.1 & 9.1±1.7 & 16.8±0.3 & \textbf{27.5±1.1} & \textbf{42.5±2.4} \\
 &  & ASM & 6.2±1.7 & 8.5±2.9 & 9.7±1.9 & 16.4±1.7 & 25.1±2.3 & 39.3±3.2 \\
\multirow{7}{*}{} & \multirow{2}{*}{} & IMP & 5.4±1.5 & 7.6±1.4 & 9.6±0.7 & 15.3±3.1 & 24.9±3.4 & 39.9±0.9 \\
 &  & ALL & 5.6±3.2 & 8.3±2.2 & 9.0±1.6 & 15.2±0.5 & 24.3±1.8 & 39.7±1.8 \\
 \midrule
 & \multirow{5}{*}{TF} & - & 17.0±1.1 & 17.0±0.8 & 19.6±0.9 & 23.1±2.0 & 26.4±0.7 & 36.0±0.8 \\
 &  & INT & \textbf{20.3±1.5} & 20.6±1.0 & 22.5±1.4 & 24.8±1.7 & \textbf{28.1±0.1} & 41.0±3.4 \\
 &  & ASM & 19.3±0.0 & 20.8±1.1 & 22.5±1.2 & \textbf{25.8±0.5} & 27.3±1.7 & 40.0±0.4 \\
 &  & IMP & 20.2±1.4 & 20.5±2.4 & 21.4±0.3 & 24.8±1.0 & 27.5±1.1 & 40.1±2.8 \\
 &  & ALL & 19.9±1.8 & \textbf{22.1±0.7} & \textbf{22.9±1.2} & 25.5±1.2 & 27.1±1.6 & \textbf{41.9±1.4} \\
 \midrule
\multirow{3}{*}{t5-base} & \multirow{3}{*}{ID} & - & 5.3±4.6 & 8.0±4.0 & 10.0±2.1 & 21.1±2.6 & 26.9±0.6 & 42.8±1.7 \\
 &  & INT & 7.1±0.2 & 8.7±4.0 & 15.1±0.9 & 18.6±1.8 & 26.3±3.1 & \textbf{45.0±0.7} \\
 &  & ASM & 8.9±2.2 & 8.1±0.8 & 10.3±5.6 & 20.3±1.1 & \textbf{28.6±0.2} & 43.1±0.6 \\
\multirow{7}{*}{} & \multirow{2}{*}{} & IMP & 6.3±1.3 & \textbf{13.5±2.1} & 18.3±2.7 & \textbf{22.8±1.7} & \textbf{28.6±1.8} & 42.0±0.8 \\
 &  & ALL & 6.5±3.8 & 12.9±3.2 & \textbf{18.4±1.7} & 22.0±2.0 & 24.8±0.8 & 44.2±1.2 \\
 \midrule
 & \multirow{5}{*}{TF} & - & 19.8±0.7 & 20.6±0.2 & 20.6±1.0 & 24.6±2.1 & 27.8±1.0 & 41.7±1.0 \\
 &  & INT & 22.3±0.7 & 22.5±0.9 & 22.3±0.6 & \textbf{27.6±0.6} & \textbf{31.6±1.7} & \textbf{43.9±0.5} \\
 &  & ASM & 21.0±0.6 & 21.3±1.2 & 21.6±1.1 & 25.4±1.2 & 28.2±1.0 & 43.5±0.6 \\
 &  & IMP & 22.4±1.0 & 21.9±0.5 & 22.5±1.3 & 25.4±1.1 & 27.9±3.7 & 40.5±2.6 \\
 &  & ALL & \textbf{23.1±0.5} & \textbf{23.2±0.3} & \textbf{23.2±0.5} & 27.1±1.9 & 29.0±1.0 & \textbf{43.9±1.6} \\
\bottomrule
\end{tabular}
\end{table*}

\begin{table*}[]
\centering
\small
\caption{Analysis of dialogue utterances with corresponding contextual information and labels when rationales (RAT) are always better.}
\resizebox{\textwidth}{!}{
\begin{tabular}{p{0.08\linewidth}|p{0.15\linewidth}|p{0.3\linewidth}|p{0.08\linewidth}|p{0.4\linewidth}|p{0.1\linewidth}}
\toprule
\textbf{Dataset} & \textbf{UTT} & \textbf{Context} & \textbf{Model} &\textbf{RAT} & \textbf{Gold Label / Pred with RAT / Pred with UTT} \\
\midrule
friends 
& Ross: ! 
& Chloe: Do I know why we're rushing? \newline Ross: Yeah, y'know the ah, the girlfriend I told you about last night? \newline Ross: Well it turns out she ah, she wants to get back together with me. \newline Ross: Oh, I found it! 
& BERT-ALL
& The speaker's intention in the final utterance is to express excitement or surprise. The exclamation mark indicates a sudden realization or discovery. \newline Based on the dialogue history, it can be assumed that Ross has found something important or significant. The exclamation suggests that he has found something he was looking for or something that has a positive impact on the situation. \newline The implicit information in the final utterance is that Ross has found something that is relevant to the previous conversation. It implies that this discovery may have an impact on his relationship or the situation he was discussing with Chloe. The content of what Ross has found is not explicitly stated but can be inferred as being significant or positive based on his exclamation.
& Surprise / Surprise / Neutral \\
\midrule
iemocap & 
F: Thanks. 
& F: Thank you \newline M: Of course. \newline M: Is there a funeral planned? or- \newline F: Yeah, they're going to do some sort of memorial service or something. \newline M: Cool, Well, If you want me to go with you, I will. 
& T5-INT 
& The speaker intends to express gratitude for the offer of support and condolences given by the person they are speaking to. \newline The speaker assumes that the person they are speaking to genuinely cares about their well-being and is offering to accompany them to the memorial service. \newline The speaker implies that they appreciate the kindness and support shown by the person they are speaking to. The final utterance also suggests that the speaker may be feeling a sense of comfort or relief knowing that they have someone who is willing to be there for them during this difficult time.
& Sadness / Sadness / Other \\
\midrule
P4G 
& Persuadee: When finished with this task I will be sure to check the website. 
& Persuader: Well it's a global non-profit charity based around helping children. \newline Persuadee: What are the donations mostly used for? \newline Persuader: They are used for helping children in a variety of areas: access to education, healthcare, protection from abuse, and relief from natural disasters or conflicts. \newline Persuadee: How are donations solicited? \newline Persuader: Well you can go on their website URL or it can be deducted from this task payment. 
& T5-ALL
& The Persuadee mentions their intention to visit the organization's website after completing the current task. \newline The Persuadee assumes that the Persuader may be interested in supporting the organization and wants to assure them that they will follow up. \newline The Persuadee expresses a commitment to checking the organization's website, indicating a potential interest in making a donation or getting involved.
& Hesitance / Hesitance / Not A Resistance Strategy \\
\midrule
CB 
& Buyer: that would be nice. but how about 180 since im the one picking it up and with its one handle missing? 
& Seller: It is very good. It is actually a family heirloom passed down by my grandmother and she recently passed away 2 days ago. The nightstand is empty however. \newline Buyer: I'm sorry to hear to that. For how much do you wanna give it away? \newline Seller: Well her will listed the value of this item at 200 US dollars, but if you pick it up from her home I am willing to go lower. \newline Buyer: How lower would you go then? \newline Seller: 190\$ and I can provide lemonade as you move it out.
 & T5-ALL 
 & The buyer suggests a further reduction to 180\$ due to the missing handle on the nightstand. \newline The buyer assumes that the missing handle affects the value of the nightstand and justifies a lower price. \newline The buyer is negotiating a lower price based on the condition of the nightstand.
 & Counter Argumentation / Counter Argumentation / Personal Choice \\

 \bottomrule
\end{tabular}}
\label{table:utterance_analysis_better}
\end{table*}

\begin{table*}[]
\centering
\small
\caption{Analysis of dialogue utterances with corresponding contextual information and labels when rationales are always worse}
\resizebox{\textwidth}{!}{
\begin{tabular}{p{0.08\linewidth}|p{0.2\linewidth}|p{0.35\linewidth}|p{0.08\linewidth}|p{0.35\linewidth}|p{0.2\linewidth}}
\toprule
\textbf{Dataset} & \textbf{UTT} & \textbf{Context} & \textbf{Model} &\textbf{RAT} & \textbf{Gold Label / Pred with RAT / Pred with UTT} \\

\midrule
friends & 
Ross: What the hell happened on that beach?! 
& Ross: Yeah I guess I'll let it go \newline Ross: But you-you understand how-how hard it is to forget about this. \newline Joey: Sure, it's hard to forget! \newline Joey: But that doesn't mean you have to talk about it! \newline Joey: A lot of things happened on that trip that we should never, ever talk about.
& BERT-ALL 
& The speaker's intention in the final utterance is to express frustration and curiosity about the events that transpired on the beach. Ross is seeking answers and wants to know the details of what happened. \newline The assumption here is that something significant or unexpected occurred on the beach, and Ross feels the need to understand and address it. The speaker assumes that there is information that has not been shared or discussed, and Ross wants to uncover the truth. \newline The implicit information in the final utterance is that there was an incident or series of events on the beach that have not been fully disclosed or explained. It suggests that the speaker believes there is a hidden story or secret that needs to be revealed for a complete understanding of the situation. Additionally, it implies that the beach incident is causing tension and unresolved issues among the characters involved.
& Surprise / Anger / Surprise \\

\midrule
iemocap 
& F: I'm just worried.
& M: If it's rent or something like that, I mean I can help you out a little bit, if- if that will help you you know stick with it.
\newline F: I don't know. Maybe I can take out a loan. I just- I need prospects and I don't feel like I have any. \newline M: Well, just don't give up. And there's- you know, You never know, something might be around the corner tomorrow. \newline F: But that doesn't help me when I'm hungry now. \newline M: Well, you know I'm not going to let you starve.
& T5-INT 
& The speaker intends to express their concern and anxiety about their current situation. \newline The speaker assumes that their worries and fears have been discussed and understood by the person they are speaking to. \newline The speaker implies that they have been discussing their difficulties and challenges with the person they are speaking to, and that their worries are related to their current circumstances. The speaker also implies a sense of vulnerability and uncertainty about the future.
& Sadness / Other / Sadness \\

\midrule
P4G 
& Persuadee: Perhaps a link to an organization or other agency that rates major charities would be more helpful. 
& Persuadee: I'm afraid for me, their reputation is still bad. \newline Persuadee: Sorry, no. \newline Persuader: URL Is there website! \newline Persuader: You can check them out. \newline Persuadee: Actually, their own website may be a biased barometer of their giving. 
& T5-ALL
& The Persuadee proposes an alternative approach by suggesting a link to an organization or agency that rates major charities. \newline The Persuadee assumes that relying on an organization or agency that rates major charities would provide a more objective and reliable assessment. \newline The Persuadee values objectivity and reliability when it comes to evaluating the subject's giving and believes that an external organization or agency can provide a more accurate assessment.
& Source Derogation / Counter Argumentation / Source Derogation strategy \\

\midrule
CB 
& Buyer: I just want to make sure they work and are quality / not deffective 
& Seller: Are you interested in the Subwoofer? It's a beauty. \newline Buyer: It looks good, but wondering a few things, how old is it? \newline Seller: I bought it six months ago, but I never actually took it out of the original box. It really has never been used. \newline Buyer: Oh, why is that? \newline Seller: I expected to have more time. I got sent on a 3 month business trip for my work and never got around it.
 & T5-ALL
 & The buyer wants to ensure that the Subwoofer is in working condition and of good quality. \newline The buyer assumes that there might be a risk of the Subwoofer being defective or of poor quality. \newline The buyer wants to protect their investment and avoid purchasing a faulty or subpar Subwoofer.
 & Source Derogation / Information Inquiry / Source Derogation \\

\bottomrule
\end{tabular}}
\label{table:utterance_analysis_worse}
\end{table*}

\end{document}